\documentclass[3p]{elsarticle}

\usepackage{amsmath,amssymb,amsfonts}
\usepackage{graphicx}
\usepackage{booktabs}
\usepackage{multirow}
\usepackage{multicol}
\usepackage{array}
\usepackage{siunitx}
\usepackage{threeparttable}
\usepackage{subcaption}
\usepackage{float}
\usepackage{algorithm}
\usepackage{algpseudocode}
\usepackage{url}
\usepackage{xcolor}
\usepackage[normalem]{ulem}

\journal{Expert Systems with Applications}

\begin{document}

\begin{frontmatter}

\title{Agentic SABRE: An Uncertainty-Aware Neuro-Symbolic Multi-Agent Framework for Adaptive Ransomware Detection}

\author[1]{Henry Kabuye}

\author[1]{Biju Issac\corref{cor1}}
\ead{bissac@ieee.org}
\cortext[cor1]{Corresponding author}

\author[1]{Jeyamohan Neera}

\address[1]{School of Computer Science,
Northumbria University, Newcastle Upon Tyne, UK}

\begin{abstract}
\label{sec-abstract}
Ransomware has evolved into a complex, adaptive, and fast–moving adversary category in which static signatures and monolithic classifiers fail to generalise under concept drift, evasion, and behavioural polymorphism. In this paper we present \textbf{Agentic SABRE (Semantic–Behavioural Arbitration for Ransomware Evaluation)}: an uncertainty–aware, neuro–symbolic, multi–agent framework for adaptive ransomware detection. SABRE fuses \emph{semantic} (representation–based) and \emph{behavioural} (time–window forensic telemetry) evidence, and employs Monte Carlo Dropout inference to quantify epistemic uncertainty for each agent.

We introduce a decision–layer orchestrator that performs risk– and uncertainty–aware triage via two interpretable thresholds: a risk score $\tau$ and an uncertainty budget $\kappa$. High–confidence, high–risk samples are automatically contained, while uncertain or borderline cases are escalated to human analysts, establishing a flexible computational contract between autonomous response and analyst oversight. To support auditability and trust, SABRE integrates post–hoc explainability mechanisms including gradient saliency, permutation importance, and counterfactual analysis, enabling both local and global interpretation of agent decisions.

Extensive evaluation on RDset and RanSMAP demonstrates that Agentic SABRE preserves perfect discrimination on saturated semantic datasets (AUC $=1.0$) while improving robustness under weak behavioural signals, achieving up to a $4.9\%$ relative reduction in false escalations at equal recall and maintaining calibrated predictive uncertainty. Counterfactual analysis further shows that semantic and behavioural decisions can be flipped with bounded perturbation cost, indicating stable and interpretable decision boundaries. Overall, Agentic SABRE is not merely a higher–accuracy detector but an agentic cyber–defence system that combines uncertainty–aware automation, explainable reasoning, and adaptive triage under evolving ransomware threats.
\end{abstract}

\end{frontmatter}

\section{Introduction}
\label{sec:introduction}

\noindent\textbf{R}ansomware has become one of the most disruptive forms of cybercrime, with modern families exhibiting rapid mutation, behavioural polymorphism, and highly adaptive strategies designed to evade both traditional signature-based defences and contemporary machine learning detectors ~\cite{fernando2020study, razaulla2023age}. These strains frequently employ advanced obfuscation, polymorphism, and metamorphism to alter operational behaviour and code signatures, undermining static detection techniques and forcing dependence on behavioural and AI-based methods ~\cite{fernando2020study, Urooj2024polymorphism}. Enterprise environments now operate under continuous concept drift, where benign and malicious behaviours evolve over time and trained models degrade in performance unless retrained or adapted ~\cite{Ceschin2022ConceptDrift}. Static classifiers that assume fixed data distributions consequently suffer rapid decay in effectiveness as attacker strategies evolve ~\cite{Fernando2024FeSAD, razaulla2023age}. These trends expose the limitations of monolithic ransomware classifiers that lack explicit representations of uncertainty, fail to adapt to temporal distributional changes without costly retraining, and provide no mechanism for calibrated human–AI interaction, a gap increasingly recognised in both malware analytics and adaptive learning research.

Machine learning has been increasingly applied to ransomware detection, ranging from classical feature-based models to deep learning approaches that operate on system call traces, filesystem activity, API semantics, and other forensic telemetry ~\cite{alraizza2023ransomware, zahoora2022Deep}. While such models can achieve strong in-distribution performance, they often behave unreliably on out-of-distribution samples, novel ransomware variants, or previously unseen system behaviours ~\cite{fernando2020study}. Crucially, standard neural detectors provide point predictions without quantifying epistemic uncertainty, making it difficult to determine when the model is confident enough to support autonomous containment or when human escalation is required ~\cite{Doan2024uncertainBay}. This absence of uncertainty-awareness leads to undesirable trade-offs: aggressive thresholds may cause catastrophic false positives, whereas conservative thresholds increase analyst workload and delay incident response ~\cite{Ispahany2024Survey}.

To address these challenges, we propose Agentic SABRE (Semantic–Behavioural Arbitration for Ransomware Evaluation), an uncertainty-aware, neuro-symbolic, multi-agent framework for adaptive ransomware detection. Rather than relying on a single model, SABRE decomposes detection into specialised \emph{semantic} and \emph{behavioural} agents, each trained on distinct but complementary signals. Both agents employ Monte Carlo Dropout to produce calibrated estimates of epistemic uncertainty, enabling the system to distinguish between high-confidence predictions and unreliable or ambiguous cases. A decision-layer orchestrator fuses agent scores and applies a triage policy based on a risk threshold \(\tau\) and an uncertainty threshold \(\kappa\),
enabling autonomous containment only when the fused prediction is both
high-risk and low-uncertainty. Uncertain or moderately risky samples are safely escalated to human analysts, creating a principled interface between automated defence and human oversight. Additionally, SABRE incorporates lightweight symbolic cues (e.g., ATT\&CK tactics, API-type ontologies, and heuristic rule signals) to support neuro-symbolic reasoning and improve robustness against model blind-spots. Whereas existing approaches quantify uncertainty to interpret predictions~\cite{kabuye2025explainable}, Agentic SABRE operationalises uncertainty to govern system actions.

\subsection{Contributions}
This work makes the following key contributions:
\begin{itemize}
    \item We introduce \textbf{Agentic SABRE}, a multi-agent architecture that
    integrates semantic embeddings and behavioural telemetry for ransomware
    detection under concept drift.

    \item We develop an \textbf{uncertainty-aware inference mechanism} using
    Monte Carlo Dropout, enabling each agent to output both a predictive mean and
    epistemic uncertainty.

    \item We propose a \textbf{triage policy} governed by a risk threshold
    \(\tau\) and uncertainty threshold \(\kappa\), enabling calibrated decisions
    between autonomous containment, escalation, and benign allowance.

    \item We incorporate \textbf{lightweight symbolic reasoning} via
    ATT\&CK-inspired heuristic cues---including API call category indicators,
    entropy-threshold signals, and ransomware-relevant filesystem access
    patterns---integrated into the feature extraction and triage stages to
    enrich agent decisions with interpretable threat-intelligence signals,
    forming a hybrid neuro-symbolic detection pipeline.

    \item Through extensive experiments across multiple ransomware families and
    benign workloads, we show that SABRE improves robustness under drift, reduces
    false escalation volume at equal recall, and maintains calibrated predictive
    uncertainty.
\end{itemize}

\subsection*{What Makes This Different from Ensemble-Based Approaches}

A key concern in multi-agent ransomware detection is whether the proposed system is simply
an ensemble in disguise. Agentic SABRE is \emph{fundamentally distinct} from conventional
ensemble methods in three respects. First, the two agents operate on \emph{structurally
heterogeneous} input spaces---static PE embedding vectors ($\mathbb{R}^{384}$) versus
sliding-window I/O telemetry statistics ($\mathbb{R}^{16}$)---not on bootstrap resamples
of the same feature space. Their roles are complementary by design, not redundant. Second,
neither agent's probability output alone determines the triage decision: the orchestrator
also conditions on the \emph{epistemic uncertainty} of each agent, which would be unavailable
in a standard ensemble voting scheme. Third, the triage policy is executable and
interpretable---it is a formal control mechanism, not a soft label aggregation---meaning
the system can unambiguously withhold autonomous action under ambiguity, a property that
vanilla ensembles do not possess. Together, these properties make Agentic SABRE an agentic
cyber-defence system rather than a higher-accuracy detector.

\subsection*{Paper structure}
The remainder of this paper is organised as follows.
Section~\ref{sec:background} provides background on ransomware detection and
motivates the need for adaptive, uncertainty-aware defence under concept drift
and adversarial evolution. Section~\ref{sec:related} reviews related work on
ransomware detection, uncertainty quantification, explainable security analytics,
and neuro-symbolic and multi-agent approaches, highlighting limitations that
motivate the proposed framework. Section~\ref{sec:method} presents the Agentic~SABRE
architecture, including the multi-agent design, uncertainty estimation via Monte
Carlo Dropout, score fusion, and the uncertainty-aware triage policy.
Section~\ref{sec:threatmodel} formalises the threat model and adversarial
assumptions. Section~\ref{sec:experimental_setup} describes the experimental
setup, including datasets, feature construction, model training, and evaluation
protocols. Section~\ref{sec:results} reports empirical results and analyses the
behaviour of the uncertainty-aware triage policy under different operating
regimes. Section~\ref{sec:explainability} presents explainability and
interpretability analyses based on feature sensitivity and counterfactual
reasoning. Section~\ref{sec:ablation} reports ablation studies examining the
contribution of key system components. Section~\ref{sec:discussion} discusses
operational implications and theoretical insights, while
Section~\ref{sec:limitations} outlines limitations and directions for future
work. Finally, Section~\ref{sec:conclusion} concludes the paper.

\section{Background}
\label{sec:background}

This section establishes the technical context for Agentic SABRE by reviewing the ransomware
threat landscape, summarising the limitations of existing detection approaches, and motivating
the design requirements addressed by the proposed framework.

Ransomware has become a dominant form of cybercrime, driven by increasingly
sophisticated techniques for evasion, polymorphism, and operational stealth.
Traditional approaches to ransomware detection have relied on signature
matching, heuristic rule sets, or simple statistical anomaly detection.
Although effective for well-known threats, such approaches degrade rapidly in the presence of obfuscation, encryption, behavioural drift, and previously unseen attack families. As enterprise systems evolve and ransomware variants diversify, detection frameworks must adapt to changing distributions rather than relying on static assumptions~\cite{fernando2020study, razaulla2023age}.

Machine learning has emerged as a powerful tool for ransomware detection,
leveraging system call traces, API telemetry, filesystem access patterns, and related forensic signals ~\cite{zahoora2022Deep, alraizza2023ransomware, Alraizza2025StorageAP}. Classical models and deep neural networks have shown promise in modelling such behavioural and semantic features. However, purely model-based detection suffers from three persistent challenges. First, data imbalance remains severe, as benign activity dominates operational telemetry. Oversampling techniques such as SMOTE and generative models such as CTGAN have been proposed to mitigate minority-class scarcity, improving training stability and representation of rare malicious events~\cite{chawla2002smote, xu2019ctgan}. Second, semantic modelling using transformers and other embedding-based architectures has improved contextual understanding of system logs and API sequences, yet these models often lack robustness under concept drift and out-of-distribution (OOD) conditions ~\cite{fernando2020study}. Third, most existing machine learning detectors provide only point predictions and do not quantify predictive uncertainty, making it difficult to determine whether a model's output is reliable enough to drive autonomous containment decisions~\cite{gal2016dropout, gawlikowski2023Uncertainity}.

Recent work in cybersecurity has begun to explore uncertainty quantification as a means of addressing model unreliability in deployment ~\cite{gal2016dropout, gawlikowski2023Uncertainity, kabuye2025explainable}. Techniques such as Monte Carlo Dropout and deep ensemble variance estimation allow detectors to express confidence levels alongside predicted probabilities. This is critical in high-stakes applications where false positives incur operational cost and false negatives cause severe damage. Nevertheless, most existing cybersecurity systems use uncertainty only as an auxiliary diagnostic tool rather than integrating it into a structured triage or escalation policy. As a result, model uncertainty is rarely incorporated into actionable decision-making and is not used to determine when to escalate suspicious activity to a human analyst.

Parallel to these developments, neuro-symbolic and multi-agent architectures have gained traction for their modularity and interpretability~\cite{Besold2021neurosymbolicsurvey}. Multi-agent designs allow distinct models to specialise in complementary aspects of behaviour, such as semantic representations versus resource-level telemetry, enabling more robust detection in complex environments. Symbolic cues—such as ATT\&CK tactics, API ontologies, and handcrafted behavioural rules—have also been used to enhance interpretability and reduce blind spots in purely neural systems. However, most prior systems treat multi-agent fusion as a simple ensemble procedure, and few integrate uncertainty-aware reasoning or explicit policies for calibrated automation.

In summary, prior research has advanced data augmentation, semantic embeddings, behavioural modelling, multi-agent architectures, and explainability techniques, but these efforts often address each challenge in isolation. Existing systems typically lack: (i) modular agents specialising in different behavioural and semantic channels, (ii) principled uncertainty quantification at the agent level, (iii) a calibrated fusion mechanism that incorporates both risk and uncertainty, and (iv) a structured triage policy linking model confidence to autonomous or human-in-the-loop decisions. Prior surveys of ransomware detection research highlight these fragmented efforts and underscore gaps in unified, operationally grounded frameworks for addressing challenges such as concept drift, uncertainty quantification, and integrated decision policies ~\cite{Ispahany2024Survey}. These gaps motivate the design of \textbf{Agentic SABRE}, a unified uncertainty-aware, neuro-symbolic, multi-agent framework explicitly tailored for adaptive ransomware detection under drift, heterogeneity, and operational constraints.

While prior surveys comprehensively catalogue ransomware detection techniques, they do not propose operational mechanisms that bind predictive uncertainty to autonomous decision-making. Agentic SABRE addresses this gap by embedding uncertainty directly into an executable triage policy.

\section{Related Work}
\label{sec:related}

This section surveys the four bodies of prior work most relevant to Agentic SABRE---ransomware
detection, data augmentation, uncertainty quantification in cybersecurity, and multi-agent
neuro-symbolic approaches---and identifies the gaps that motivate our design.

Research on ransomware detection intersects multiple areas including behavioural modelling, semantic reasoning, uncertainty-aware machine learning, and modular multi-agent architectures. We review the most relevant lines of work and highlight the distinctions between existing approaches and the design of Agentic~SABRE. Prior work on uncertainty-aware ransomware and intrusion detection typically treats uncertainty as a diagnostic signal: uncertainty estimates are reported to aid interpretation, calibration analysis, or analyst awareness, but they do not explicitly constrain system behaviour. In such settings, uncertainty remains an auxiliary output rather than a governing variable in the decision process. In contrast, Agentic SABRE treats epistemic uncertainty as an operational decision variable. Uncertainty is explicitly incorporated into the triage policy and directly determines whether the system may autonomously contain a process, escalate to human analysts, or allow execution. This transforms uncertainty from a descriptive property of predictions into a prescriptive control signal that enforces safety and conservatism under ambiguity.

\subsection{Machine Learning for Ransomware Detection}

Behavioural ransomware detectors traditionally rely on handcrafted indicators such as entropy, API counts, I/O throughput, or memory access volatility. These features have been used in conjunction with classical machine learning models such as SVMs and random forests, but such approaches struggle to generalise to zero-day variants and polymorphic families. Recent deep learning systems extend this line of work by leveraging system calls, registry traces, and log sequences. Transformer-based architectures~\cite{vaswani2017attention} and LLM-derived embeddings have also been explored for intrusion detection and semantic log analysis~\cite{tian2025llmsurvey, BenchmarkingLLMsPhishing2024,
CANAL_CTILLM2024}. However, these systems generally operate as monolithic
detectors and produce deterministic outputs without quantifying epistemic
uncertainty or providing structured mechanisms for human escalation. In contrast, Agentic~SABRE incorporates per-agent Monte Carlo Dropout uncertainty estimation, enabling calibrated confidence-aware triage. Recent work has also explored deep learning and blockchain-based frameworks for malware detection in IoT environments~\cite{pawar2024securing}, underscoring the breadth of architectural approaches being applied to adaptive threat analysis and explainable malware defence.

\subsection{Data Augmentation and Imbalance Handling}

Imbalanced datasets remain a pervasive challenge in ransomware detection because benign activity vastly outweighs malicious traces. Oversampling approaches such as SMOTE~\cite{chawla2002smote} and generative tabular models such as CTGAN ~\cite{xu2019ctgan, ashrapov2020tabular} have been widely adopted to mitigate class imbalance. Such approaches have proven effective in cybersecurity domains, including IoT botnet detection~\cite{HabibiCL23} and enterprise anomaly detection~\cite{AIenhancedDefense2022}. While these methods improve training
stability, they are typically employed only at the data preprocessing stage and are rarely integrated into multi-stage detection pipelines. Agentic~SABRE differs by applying CTGAN not to raw features but to the \emph{score space}, enriching rare high-risk cases entering the fusion layer without altering agent-level probabilities or uncertainties.

\subsection{Semantic Modelling and Log Understanding}

LLMs and transformer-based encoders have demonstrated strong performance in
security log analysis and anomaly detection~\cite{tian2025llmsurvey, BenchmarkingLLMsPhishing2024}. Systems such as LLM-LADE~\cite{ZhangLZYHY25}
combine BERT-style embeddings with SHAP explanations~\cite{lundberg2017shap} to provide interpretable semantic anomaly detection. However, these systems focus on general log anomalies and do not exploit structured behavioural indicators (e.g., entropy, GPA variance, throughput) that are crucial in ransomware analysis. Furthermore, prior semantic models typically operate as standalone classifiers and do not integrate uncertainty-aware fusion or symbolic security knowledge. SABRE diverges from these approaches by combining semantic embeddings with
behavioural telemetry in a modular multi-agent design enriched with symbolic cues from API ontologies and ATT\&CK-aligned heuristics.

\subsection{Uncertainty Quantification in Cybersecurity}

There is growing interest in incorporating uncertainty estimation into deep
learning models for security~\cite{kabuye2025explainable}. Methods such as Monte Carlo Dropout, deep ensembles, and Bayesian neural networks have been used to improve robustness and identify unreliable predictions. However, existing systems typically use uncertainty as an auxiliary diagnostic rather than as a structured driver of decision policies. Few works integrate uncertainty thresholds into automated response pipelines or calibrate model behaviour using formal triage criteria. Agentic~SABRE advances this direction by embedding uncertainty estimates directly into an operational triage policy governed by risk and uncertainty thresholds (\(\tau\), \(\kappa\)), enabling selective automation with safety guarantees.

\subsection{Multi-Agent and Neuro-Symbolic Approaches}

Multi-agent learning has been explored for adaptive intrusion detection,
cooperative anomaly detection, and adversarial resilience~\cite{AdversarialCooperationLLM2024, MultiAgentAdaptiveDetection2023}. Parallel advances in neuro-symbolic systems highlight the benefits of combining neural representations with symbolic rules and knowledge bases~\cite{YangSBK25}. Explainable deep models optimised through meta-learning and automated tuning~\cite{FaresE25} further demonstrate the value of modular, interpretable architectures in cybersecurity.

A critical distinguishing feature of Agentic SABRE relative to existing neuro-symbolic cybersecurity frameworks concerns the location and mutability of the symbolic component. Prior neuro-symbolic systems---including the knowledge-base integration approaches surveyed by Yang et al.~\cite{YangSBK25} and the neural-symbolic architectures reviewed by Besold et al.~\cite{Besold2021neurosymbolicsurvey}---embed symbolic rules at the architecture or knowledge-base level, meaning that the symbolic and neural components are structurally coupled and cannot be updated independently. In Agentic SABRE, by contrast, the symbolic component takes the form of ATT\&CK-inspired interpretable triage thresholds: API call category flags, entropy-threshold signals, and filesystem access pattern heuristics that are evaluated by the orchestrator as conditions on the fused risk score and epistemic uncertainty. These symbolic cues are encoded as threshold parameters ($\tau$, $\kappa$, $\tau_{\text{high}}$, $\kappa_{\text{low}}$) that can be recalibrated or updated independently of the underlying neural agents---without retraining---whenever threat intelligence changes or deployment conditions shift. This architectural separation between the learning layer and the symbolic reasoning layer is a deliberate design choice that enables policy-level agility while preserving agent-level discriminative capacity.

However, these prior systems typically lack (i) agent-level uncertainty
estimation, (ii) uncertainty-aware fusion, and (iii) operational triage policies linking uncertainty and risk to automated or human-in-the-loop decisions. Agentic SABRE directly addresses all three gaps: it assigns dedicated uncertainty estimation to each agent, fuses both risk and uncertainty into a calibrated decision signal, and enforces an executable triage policy that determines when the system may act autonomously and when human oversight is required.

Table~\ref{tab:related_comparison} summarises how representative prior systems compare to
Agentic SABRE across the dimensions that define the design space.

\begin{table*}[t]
\centering
\caption{Comparison of Agentic SABRE with representative prior ransomware and intrusion
detection systems across key design dimensions. \checkmark~= present; $\circ$~= partial;
$\times$~= absent.}
\label{tab:related_comparison}
\resizebox{\linewidth}{!}{
\begin{tabular}{lccccccc}
\toprule
\textbf{System} &
\textbf{Multi-Agent} &
\textbf{Uncertainty} &
\textbf{Triage Policy} &
\textbf{Symbolic Cues} &
\textbf{XAI} &
\textbf{Concept Drift} &
\textbf{Score-Level Aug.} \\
\midrule
Fernando \emph{et al.}~\cite{fernando2020study}        & $\times$ & $\times$ & $\times$ & $\times$ & $\times$ & $\times$ & $\times$ \\
Razaulla \emph{et al.}~\cite{razaulla2023age}          & $\times$ & $\times$ & $\times$ & $\times$ & $\times$ & $\circ$  & $\times$ \\
Deep-learning detectors~\cite{zahoora2022Deep}         & $\times$ & $\times$ & $\times$ & $\times$ & $\circ$  & $\times$ & $\times$ \\
LLM-LADE~\cite{ZhangLZYHY25}                          & $\times$ & $\times$ & $\times$ & $\circ$  & \checkmark & $\times$ & $\times$ \\
Kabuye \emph{et al.}~\cite{kabuye2025explainable}      & $\times$ & \checkmark & $\times$ & $\times$ & \checkmark & $\times$ & $\times$ \\
Soltani \emph{et al.}~\cite{MultiAgentAdaptiveDetection2023} & \checkmark & $\times$ & $\times$ & $\times$ & $\times$ & $\circ$ & $\times$ \\
\midrule
\textbf{Agentic SABRE (ours)} & \checkmark & \checkmark & \checkmark & \checkmark & \checkmark & \checkmark & \checkmark \\
\bottomrule
\end{tabular}}
\end{table*}

\section{Agentic SABRE Framework}
\label{sec:method}

This section presents the complete Agentic SABRE architecture. We describe the five-stage
operational pipeline, the score-fusion mechanism, the uncertainty aggregation strategy, and
the design of the symbolic reasoning components. The triage policy itself is detailed in
Section~\ref{sec:uncert_policy}.

\subsection{System Overview}
\label{subsec:SysOverview}
\begin{figure*}[t]
  \centering
  \resizebox{\textwidth}{!}{%

  \includegraphics[width=\linewidth]{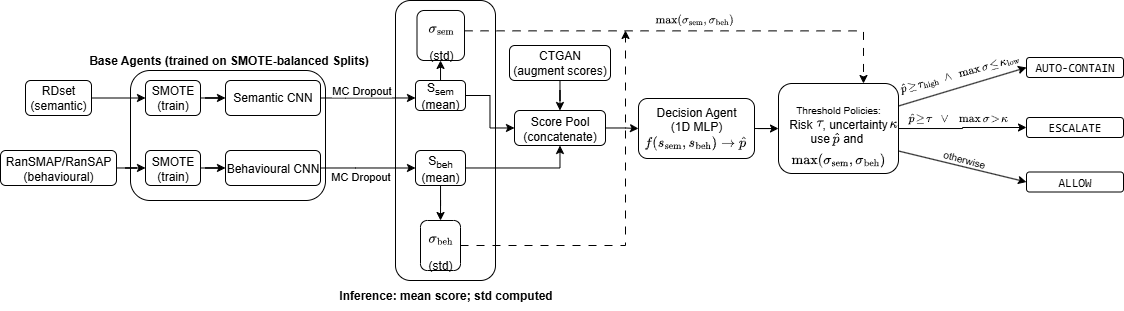}
}
  \caption{Operational pipeline with uncertainty-aware, agentic policy. Semantic and behavioural CNNs (with MC Dropout) yield mean scores and per-agent uncertainties. Mean scores are concatenated and CTGAN-augmented for the Decision Agent to produce a fused risk $\hat{p}$. A policy uses risk threshold $\tau$ and uncertainty threshold $\kappa$ (with optional stricter $\tau_{\text{high}}$, $\kappa_{\text{low}}$) to branch to \emph{AUTO-CONTAIN}, \emph{ESCALATE}, or \emph{ALLOW}.}.
 
  \label{fig:agentic-Sabre_policy-pipeline}
\end{figure*}

Agentic~SABRE is a neuro-symbolic, uncertainty-aware multi-agent system designed to detect ransomware under concept drift and behavioural polymorphism. The framework consists of five stages: (i) hybrid feature extraction, (ii) augmentation, (iii) agent training with MC Dropout, (iv) score fusion, and (v) uncertainty-aware triage. The operational pipeline is illustrated in Fig.~\ref{fig:agentic-Sabre_policy-pipeline}. Unlike conventional uncertainty-aware detectors, which expose predictive uncertainty primarily for post hoc interpretation, the proposed framework embeds uncertainty directly into the decision logic. The thresholds $\kappa$ and $\kappa_{\text{low}}$ act as explicit uncertainty budgets, defining regions in which automated action is permitted or prohibited. As a result, uncertainty is not merely observed but reasoned over: high-risk predictions with elevated epistemic uncertainty are deliberately prevented from triggering autonomous containment and are instead routed to human analysts. The modular structure of Agentic~SABRE allows the triage policy to be deployed independently of the underlying detection models, enabling policy-level updates and operational recalibration without retraining the base agents.

\subsection{Score Fusion and Decision Agent}

This subsection describes how the two agents' outputs are combined into a single ransomware risk estimate.
Let $\bar{p}_{\text{sem}} \in [0,1]$ denote the predictive mean probability of the Semantic Agent
and $\bar{p}_{\text{beh}} \in [0,1]$ that of the Behavioural Agent, each obtained via Monte Carlo
Dropout (Algorithm~\ref{alg:mc_dropout}).
The corresponding per-agent epistemic uncertainties are $\sigma_{\text{sem}}$ and $\sigma_{\text{beh}}$.
The fused score vector is constructed as:

\begin{equation}
\mathbf{s} =
\begin{bmatrix}
\bar{p}_{\text{sem}} \\
\bar{p}_{\text{beh}}
\end{bmatrix},
\label{eq:scorevec_basic}
\end{equation}

optionally extended with CTGAN-generated synthetic score samples to address score-space class imbalance.
A lightweight MLP-based Decision Agent learns a mapping from $\mathbf{s}$ and produces a unified
ransomware risk estimate:

\begin{equation}
\hat{p} = f(\mathbf{s})
\label{eq:scorepool}
\end{equation}

where $\hat{p} \in [0,1]$ is the fused risk probability used by the triage orchestrator.
Note that at deployment the full score vector additionally incorporates the per-agent uncertainties
(see Eq.~\ref{eq:scorevector} in Section~\ref{sec:results}), which are used as inputs to the
trained Decision Agent to produce the calibrated fused risk.

\subsection{Uncertainty Aggregation}

To ensure conservative behaviour, agent-level uncertainties are aggregated via a max-operator:

\begin{equation}
\sigma_{\max} = \max(\sigma_{\text{sem}}, \sigma_{\text{beh}}),
\label{eq:sigmamax}
\end{equation}

where $\sigma_{\text{sem}}$ and $\sigma_{\text{beh}}$ are the Monte Carlo Dropout standard
deviations of the semantic and behavioural agents respectively. Using the maximum rather than
the mean ensures that elevated uncertainty in \emph{either} modality---not just their
average---constrains autonomous action, providing a conservative safety guarantee.
This definition is carried consistently throughout the triage logic and fusion analysis.
\newline

\noindent\textbf{Choice of uncertainty estimator.}
Monte Carlo (MC) Dropout~\cite{gal2016dropout} was selected over deep ensembles and Bayesian
neural networks for three practical reasons. First, MC Dropout requires no architectural
changes beyond retaining dropout at inference time, making it straightforward to integrate
into the existing CNN agents. Second, its per-sample inference cost is $T$ stochastic forward
passes ($T = 30$ in our experiments), which is lower than training and storing $M$ independent
ensemble members ($M \ge 5$ is typical). Third, the agents' training pipelines already include
dropout regularisation, so no additional training overhead is incurred. We acknowledge that
deep ensembles can produce better-calibrated uncertainty under strong distribution shift,
as they approximate a wider posterior over model weights. MC Dropout may under-estimate
epistemic uncertainty in highly non-stationary behavioural regimes---a limitation noted in
Section~\ref{sec:limitations} and left as a direction for future work with ensemble-based
SABRE variants.

\section{Uncertainty-Aware Triage Policy}
\label{sec:uncert_policy}

This section formalises the triage decision mechanism that converts the fused risk score
$\hat{p}$ and aggregated uncertainty $\sigma_{\max}$ into one of three operational actions:
\textsc{AUTO-CONTAIN}, \textsc{ESCALATE}, or \textsc{ALLOW}. We present the policy design
rationale, the decision rules, and an operational interpretation of the threshold geometry.

Agentic~SABRE can be interpreted as a decision-theoretic control system in which
epistemic uncertainty constrains autonomous action. Rather than optimising
predictive accuracy alone, the framework explicitly optimises operational utility
by regulating when automated containment is permitted and when human oversight
is required. In this formulation, epistemic uncertainty is not treated as a
diagnostic by-product of prediction, but as an explicit \emph{control signal}
that governs whether the system may act autonomously or must defer to analysts.

Given a fused ransomware risk score $\hat{p} \in [0,1]$ and an aggregated epistemic
uncertainty $\sigma_{\max} \in [0,1]$, the orchestrator applies a two-threshold
policy regulated by: (i) a risk threshold $\tau$ and (ii) an uncertainty budget
$\kappa$. For safer autonomous containment, stricter thresholds
$\tau_{\text{high}}$ and $\kappa_{\text{low}}$ define a high-confidence operating
region.

\subsection{Policy Design and Thresholds}
\label{subsec:policy_thresholds}
Axis-aligned thresholds are deliberately chosen over learned non-linear decision boundaries to preserve interpretability, auditability, and policy stability under distribution shift—properties that are critical in regulated operational environments.

The policy uses two pairs of thresholds:
(i) $(\tau,\kappa)$ defining the general escalation boundary, and
(ii) $(\tau_{\text{high}},\kappa_{\text{low}})$ defining a stricter region in which
autonomous containment is permitted.
These pairs satisfy the ordering constraints
\begin{equation}
\tau \le \tau_{\text{high}}, \qquad \kappa_{\text{low}} \le \kappa,
\label{eq:threshold_order}
\end{equation}
meaning $\tau_{\text{high}}$ demands a higher fused risk for autonomous action than $\tau$
does for escalation, and $\kappa_{\text{low}}$ imposes a tighter uncertainty ceiling than
$\kappa$. Intuitively, $\tau$ controls the minimum risk level that warrants any intervention,
while $\kappa$ constrains how much epistemic uncertainty is tolerated before human escalation
is required. The stricter pair $(\tau_{\text{high}},\kappa_{\text{low}})$ enforces a
conservative automation contract: autonomous containment is permitted only when the fused
risk is very high \emph{and} uncertainty is sufficiently low, ruling out overconfident
automated responses near the decision boundary.

\subsection{Decision Rules}
\label{subsec:policy_rules}

Let $\sigma_{\max}$ denote the aggregated epistemic uncertainty (e.g., the maximum
uncertainty across the semantic and behavioural agents). The orchestrator maps each
sample to one of three actions: \textsc{ALLOW}, \textsc{ESCALATE}, or
\textsc{AUTO-CONTAIN}, according to:

\begin{align}
AUTO-CONTAIN \quad &\text{if} \quad   \hat{p}\ge \tau_{\text{high}}\;\wedge\;\sigma_{\max}\le \kappa_{\text{low}}, \label{eq:policy_auto} \\
ESCALATE \quad &\text{if} \quad  \hat{p}\ge \tau \;\;\vee\;\; \sigma_{\max} > \kappa, 
\label{eq:policy_escalate}  \\
ALLOW \quad &\text{otherwise}. && \label{eq:policy_allow}
\end{align}

\subsection{Operational Interpretation}
\label{subsec:policy_interpretation}

The policy defines a calibrated balance between autonomous defence and analyst workload.
High-risk, low-uncertainty samples fall into the \textbf{AUTO-CONTAIN} region
(Eq.~\ref{eq:policy_auto}), enabling immediate response when the system is confident.
Samples that are either sufficiently risky or epistemically uncertain are routed to
\textbf{ESCALATE} (Eq.~\ref{eq:policy_escalate}), preventing overconfident automation
under distribution shift, concept drift, or ambiguous evidence.
Remaining samples are \textbf{ALLOW} (Eq.~\ref{eq:policy_allow}), corresponding to
low-risk decisions made with acceptable confidence.

From a deployment perspective, $\tau$ and $\kappa$ act as user-tunable controls that
trade off false escalations versus false allowances, while
$(\tau_{\text{high}},\kappa_{\text{low}})$ provides an additional safety margin that
limits automated containment to the highest-confidence cases.

Algorithms~\ref{alg:agentic_sabre_pipeline}--\ref{alg:triage_policy}
collectively specify the training, inference, score fusion, and deployment-time
decision logic of Agentic~SABRE.

\begin{algorithm}[t]
\caption{Agentic SABRE: Training and Deployment Pipeline}
\label{alg:agentic_sabre_pipeline}
\begin{algorithmic}[1]
    \State \textbf{Input (training):} Raw semantic data $D_{\text{sem}}$, behavioural logs $D_{\text{beh}}$
    \State \textbf{Input (deployment):} New sample $x_{\text{sem}}, x_{\text{beh}}$
    \State \textbf{Output (deployment):} Decision $\in \{\text{AUTO-CONTAIN}, \text{ESCALATE}, \text{ALLOW}\}$
    \vspace{1mm}
    \Statex \textbf{// Phase 1: Hybrid feature extraction}
    \State Extract semantic embeddings $\mathbf{z}_i$ from $D_{\text{sem}}$ using encoder $\mathcal{E}(\cdot)$
    \State Extract behavioural feature vectors $\mathbf{b}_i$ from $D_{\text{beh}}$ via sliding-window statistics
    \vspace{1mm}
    \Statex \textbf{// Phase 2: Feature-level augmentation}
    \State Apply SMOTE to $\{\mathbf{z}_i\}$ and $\{\mathbf{b}_i\}$ independently to address class imbalance
    \vspace{1mm}
    \Statex \textbf{// Phase 3: Train base agents with MC Dropout}
    \State Train Semantic Agent CNN $A_{\text{sem}}$ on $\{\mathbf{z}_i\}$
    \State Train Behavioural Agent CNN $A_{\text{beh}}$ on $\{\mathbf{b}_i\}$
    \vspace{1mm}
    \Statex \textbf{// Phase 4: Score-level augmentation and Decision Agent}
    \For{each training sample $i$}
        \State Compute mean scores $\bar{p}_{\text{sem}}^{(i)}$, $\bar{p}_{\text{beh}}^{(i)}$ via MC Dropout (Alg.~\ref{alg:mc_dropout})
        \State Form score vector $s_i = [\bar{p}_{\text{sem}}^{(i)}, \bar{p}_{\text{beh}}^{(i)}]$
    \EndFor
    \State Train CTGAN on $\{(s_i, y_i)\}$ and obtain augmented scores $\tilde{\mathcal{D}}_S$ (Alg.~\ref{alg:ctgan_scores})
    \State Train Decision Agent $A_{\text{dec}}$ on $\mathcal{D}_S \cup \tilde{\mathcal{D}}_S$
    \vspace{1mm}
    \Statex \textbf{// Phase 5: Deployment-time inference and triage}
    \State Obtain $(\bar{p}_{\text{sem}}, \sigma_{\text{sem}})$ for $x_{\text{sem}}$ via MC Dropout
    \State Obtain $(\bar{p}_{\text{beh}}, \sigma_{\text{beh}})$ for $x_{\text{beh}}$ via MC Dropout
    \State Form $s = [\bar{p}_{\text{sem}}, \bar{p}_{\text{beh}}]$ and compute fused risk $\hat{p} = A_{\text{dec}}(s)$
    \State Compute aggregated uncertainty $\sigma_{\max} = \max(\sigma_{\text{sem}}, \sigma_{\text{beh}})$
    \State Apply triage policy using thresholds $(\tau, \kappa, \tau_{\text{high}}, \kappa_{\text{low}})$ (Alg.~\ref{alg:triage_policy})
\end{algorithmic}
\end{algorithm}

\begin{algorithm}[t]
\caption{MC Dropout Uncertainty Estimation for a Neural Agent}
\label{alg:mc_dropout}
\begin{algorithmic}[1]
    \State \textbf{Input:} Trained neural agent $A$ with dropout layers, sample $x$, number of passes $T$
    \State \textbf{Output:} Predictive mean $\bar{p}$, epistemic uncertainty $\sigma$
    \vspace{1mm}
    \State Set $A$ to evaluation mode
    \State Activate dropout layers for inference (e.g., set dropout modules to \texttt{train()} only)
    \For{$t = 1$ to $T$}
        \State $p_t \gets A(x)$ \Comment{stochastic forward pass with dropout}
    \EndFor
    \State Compute predictive mean:
    \[
        \bar{p} = \frac{1}{T}\sum_{t=1}^{T} p_t
    \]
    \State Compute predictive standard deviation:
    \[
        \sigma = \sqrt{\frac{1}{T}\sum_{t=1}^{T} (p_t - \bar{p})^2}
    \]
    \State \Return $(\bar{p}, \sigma)$
\end{algorithmic}
\end{algorithm}

\begin{algorithm}[t]
\caption{Score-Level CTGAN Augmentation for Decision Agent Training}
\label{alg:ctgan_scores}
\begin{algorithmic}[1]
    \State \textbf{Input:} Trained base agents $A_{\text{sem}}, A_{\text{beh}}$; training features $\{\mathbf{z}_i\}, \{\mathbf{b}_i\}$; labels $\{y_i\}$
    \State \textbf{Output:} Augmented score dataset $\tilde{\mathcal{D}}_S$, for training $A_{\text{dec}}$
    \vspace{1mm}
    \Statex \textbf{// Step 1: Collect score tuples}
    \For{each training sample $i$}
        \State Compute $(\bar{p}_{\text{sem}}^{(i)}, \sigma_{\text{sem}}^{(i)})$ via Alg.~\ref{alg:mc_dropout} on $\mathbf{z}_i$
        \State Compute $(\bar{p}_{\text{beh}}^{(i)}, \sigma_{\text{beh}}^{(i)})$ via Alg.~\ref{alg:mc_dropout} on $\mathbf{b}_i$
        \State Form score vector $s_i = [\bar{p}_{\text{sem}}^{(i)}, \bar{p}_{\text{beh}}^{(i)}]$
    \EndFor
    \State Construct score dataset $\mathcal{D}_S = \{(s_i, y_i)\}$
    \vspace{1mm}
    \Statex \textbf{// Step 2: Train CTGAN in score space}
    \State Train CTGAN generator $G$ and discriminator $D$ on $\mathcal{D}_S$ conditioned on labels $y_i$
    \State Generate synthetic score samples $\tilde{s}_j = G(z_j, y_j)$ with noise $z_j$ and labels $y_j$
    \State Form augmented set $\tilde{\mathcal{D}}_S = \{(\tilde{s}_j, y_j)\}$
    \Statex \textbf{// Step 3: Train Decision Agent}
    \State Train Decision Agent $A_{\text{dec}}$ on $\mathcal{D}_S \cup \tilde{\mathcal{D}}_S$
    \State \Return $\tilde{\mathcal{D}}_S$
\end{algorithmic}
\end{algorithm}

\begin{algorithm}[t]
\caption{Uncertainty-Aware Triage Policy}
\label{alg:triage_policy}
\begin{algorithmic}[1]
    \State \textbf{Input:} Fused risk $\hat{p}$, agent uncertainties $\sigma_{\text{sem}}, \sigma_{\text{beh}}$
    \State \textbf{Input:} Risk thresholds $\tau$, $\tau_{\text{high}}$; uncertainty thresholds $\kappa$, $\kappa_{\text{low}}$
    \State \textbf{Output:} Decision $\in \{\text{AUTO-CONTAIN}, \text{ESCALATE}, \text{ALLOW}\}$
    \vspace{1mm}
    \State Compute aggregated uncertainty $\sigma_{\max} \gets \max(\sigma_{\text{sem}}, \sigma_{\text{beh}})$
    \vspace{1mm}
    \If{$\hat{p} \ge \tau_{\text{high}}$ \textbf{and} $\sigma_{\max} \le \kappa_{\text{low}}$}
        \State \Return \textbf{AUTO-CONTAIN} \Comment{high risk, low uncertainty}
    \ElsIf{$\hat{p} \ge \tau$ \textbf{or} $\sigma_{\max} > \kappa$}
        \State \Return \textbf{ESCALATE} \Comment{risky or uncertain}
    \Else
        \State \Return \textbf{ALLOW} \Comment{benign and confident}
    \EndIf
\end{algorithmic}
\end{algorithm}

\section{Policy Analysis}
\label{sec:policydiscussion}

This section analyses the geometry of the triage decision space, characterises the adversarial
stability of the policy boundaries, and introduces the threshold-optimisation objective used
to select operational parameters.

The triage mechanism in Agentic~SABRE is governed by the joint behaviour of the fused probability \(\hat{p}\) and the maximal epistemic uncertainty \(\sigma_{\max}\). The policy partitions the space
\begin{equation}
\mathcal{X} = \bigl\{ (\hat{p},\,\sigma_{\max}) \in [0,1] \times [0,1] \bigr\}
\label{eq:policy_space}
\end{equation}
into three disjoint regions corresponding to the actions \(\mathsf{ALLOW}\), \(\mathsf{ESCALATE}\), and \(\mathsf{AUTO\_CONTAIN}\). The decision boundaries are defined implicitly by the inequalities
\begin{equation}
\hat{p} \ge \tau_{\mathrm{high}}, \qquad \sigma_{\max} \le \kappa_{\mathrm{low}},
\label{eq:auto_rule}
\end{equation}
for autonomous containment, and
\begin{equation}
\hat{p} \ge \tau \quad \text{or} \quad \sigma_{\max} > \kappa,
\label{eq:esc_rule}
\end{equation}
for escalation. The remaining region, defined by the intersection of the complements of Eqs.~\eqref{eq:auto_rule} and \eqref{eq:esc_rule}, yields the allow decision.

The geometry of the decision space is determined by the relative magnitudes of \(\tau,\tau_{\mathrm{high}},\kappa,\kappa_{\mathrm{low}}\). The boundaries \(\hat{p}=\tau\) and \(\hat{p}=\tau_{\mathrm{high}}\) form vertical lines, while \(\sigma_{\max}=\kappa\) and \(\sigma_{\max}=\kappa_{\mathrm{low}}\) form horizontal lines. The intersection of these lines generates a partitioned rectangle whose subregions determine the triage outcome. Samples located near the boundary points,
\begin{equation}
(\tau_{\mathrm{high}}, \kappa_{\mathrm{low}}), \qquad (\tau, \kappa),
\label{eq:corners}
\end{equation}
exhibit the greatest instability under adversarial perturbation. However, due to the monotonic structure of the fusion classifier, the gradient of \(\hat{p}\) is typically largest where semantic evidence dominates, meaning the neighbourhood of \((\tau_{\mathrm{high}},\kappa_{\mathrm{low}})\) tends to be sparsely populated.

The optimisation of threshold pairs is governed by a scalarised objective that trades off
classification quality against escalation volume. Specifically, we define
\begin{equation}
J(\tau,\kappa) = \mathrm{bal\_acc}(\tau,\kappa) - \lambda\,\mathrm{EscRate}(\tau,\kappa),
\label{eq:objective}
\end{equation}
where $\mathrm{bal\_acc}(\tau,\kappa)$ is balanced accuracy under the policy,
$\mathrm{EscRate}(\tau,\kappa)$ is the fraction of samples escalated to analysts, and
$\lambda \ge 0$ controls the operational cost penalty.
This objective identifies threshold pairs that shift the decision boundaries in
Eqs.~\eqref{eq:auto_rule}--\eqref{eq:esc_rule}. Increasing \(\tau\) reduces the size of the
escalation region but simultaneously risks enlarging the allow region unless offset by a
simultaneous adjustment to \(\kappa\). The calibrated trade-off therefore emerges from solving
the non-convex problem of maximising \(J(\tau,\kappa)\) while maintaining robustness to the
uncertainty geometry defined in Eq.~\eqref{eq:policy_space}. This interplay between fused risk,
epistemic uncertainty, and threshold selection constitutes the mathematical core of the Agentic
SABRE policy framework.

\section{Threat Model}
\label{sec:threatmodel}

This section defines the adversarial assumptions under which Agentic SABRE is evaluated,
specifying what the attacker can and cannot manipulate, and explaining how the uncertainty-aware
triage policy provides robustness against the considered threat classes.

The operational setting assumes an adversary capable of deploying ransomware samples that vary in semantic structure, behavioural dynamics, or both. The defender observes two forms of telemetry: a semantic representation \(\mathbf{z}_i \in \mathbb{R}^{d_s}\) encoded from static artefacts and a behavioural representation \(\mathbf{b}_i \in \mathbb{R}^{d_b}\) extracted from temporal I/O activity. The adversary is assumed to operate under the constraint that behavioural logs and static metadata cannot be forged arbitrarily without compromising the malware’s execution.  

The threat model considers adversarial manoeuvres that attempt to reduce the classifier’s confidence while preserving functional attack behaviour. These manoeuvres can target the semantic agent by modifying static descriptors; however, such modifications typically shift the semantic embedding distribution, increasing the Wasserstein distance in Eq.~\eqref{eq:wass_sem} only marginally, and often increasing epistemic uncertainty. Similarly, behavioural mimicry attacks attempt to disguise malicious activity so that the derived empirical distributions
\begin{equation}
p(\mathbf{b}\mid y=1), \qquad  p(\mathbf{b}\mid y=0),
\label{eq:beh_dists}
\end{equation}
appear closer than reflected in Eq.~\eqref{eq:wass_beh}. Such mimicry produces elevated uncertainty in the behavioural agent due to the mismatch between expected and observed feature variance.

The model assumes that the attacker does not possess knowledge of the triage parameters \(\tau\), \(\kappa\), \(\tau_{\mathrm{high}}\), and \(\kappa_{\mathrm{low}}\), nor the calibration temperature applied to the fusion classifier. Even if such knowledge were available, the attacker would still face the constraint that reducing the fused probability \(\hat{p}\) of Eq.~\eqref{eq:fusedp} necessarily increases behavioural deviation, thus raising the uncertainty term \(\sigma_{\max}\) in Eq.~\eqref{eq:sigmamax}. Consequently, the decision boundary remains robust against a broad class of evasion strategies, since ambiguous or adversarial samples are redirected to human analysts rather than executed automatically. This model aligns with realistic constraints in ransomware operations and reflects practical adversarial limits encountered in malware obfuscation.

\section{Experimental Setup}
\label{sec:experimental_setup}

This section describes the datasets, feature construction procedures, model training
configurations, and evaluation metrics used to assess Agentic SABRE. Crucially, the two
datasets cover complementary modalities---static PE semantics and runtime behavioural
telemetry---and are never mixed at the raw feature level; fusion occurs only at the
score level, as described in Section~\ref{sec:method}.

\subsection{Datasets}
\label{subsec:datasets}

Experiments are conducted on two publicly available ransomware datasets that
exhibit complementary characteristics. RDset provides static semantic metadata derived from Portable Executable (PE) files and is used to evaluate semantic discriminability under saturated conditions. RanSMAP (and RanSAP) provide behavioural telemetry extracted from system-level I/O activity, capturing temporal variance and entropy-driven signals representative of runtime behaviour. All datasets are split into disjoint training and test sets at the sample level, ensuring that no identical execution traces or binaries appear in both splits.

RanSAP~\cite{hirano2022ransap} is a hypervisor-level behavioural dataset consisting of
low-level storage access patterns only, comprising approximately 200 million I/O events aggregated into fixed-length temporal windows collected from multiple ransomware and benign executions. RanSMAP~\cite{hirano2025ransmap}, a successor dataset, extends RanSAP by incorporating both storage and memory access patterns, yielding 11,820 execution traces across diverse hardware configurations, ransomware variants, and simultaneous benign--malicious workloads. In contrast, the dataset introduced by Mathur~\cite{mathur2020ransomware}, commonly referred to as \emph{RDset} in subsequent literature~\cite{kabuye2025explainable}, comprises 138,047 PE samples described by 57 structured static attributes. In this work, these attributes are transformed into semantic representations via a lightweight text-based encoding and embedding procedure, enabling comparison with behavioural telemetry at the agent level.

\noindent\textbf{Cross-source fusion rationale.}
RDset and RanSMAP do \emph{not} contain overlapping malware samples: RDset consists of PE
file metadata from Windows executables, while RanSMAP contains runtime I/O traces recorded
during execution on instrumented hypervisors. The two datasets are therefore drawn from
complementary observation planes of the same threat category. Raw features are
\emph{never} merged: the Semantic Agent trains exclusively on RDset embeddings and the
Behavioural Agent trains exclusively on RanSMAP telemetry. Fusion occurs only at the
\emph{score level}---the two agents' posterior mean probabilities are concatenated into the
score vector $\mathbf{s}$ (Eq.~\ref{eq:scorevec_basic}) and passed to the Decision Agent.
This design mirrors realistic deployment scenarios in which a SOC may simultaneously receive
static file analysis from a sandbox and runtime telemetry from an endpoint sensor, where
neither source alone provides complete ransomware characterisation.

\noindent\textbf{CTGAN labels.}
The CTGAN generator in Algorithm~\ref{alg:ctgan_scores} is conditioned on binary labels
$y_i \in \{0,1\}$ (benign / ransomware) drawn directly from the training set labels. No
additional label transformations or soft-label schemes are applied; the generator learns to
produce realistic score-space samples for each class independently.

\noindent\textbf{Dataset characteristics, imbalance, and contamination.}
RDset comprises 138,047 PE samples (41,123 ransomware, 96,924 benign), exhibiting heavy class imbalance of approximately 30\% ransomware. The dataset covers ransomware families active between 2014 and 2020. No temporal split is applied in the current evaluation: samples are partitioned at the binary level, so families present in the training set may appear under different instances in the test set---introducing the risk of structural similarity bias. No sample-level contamination is present: training and test splits are disjoint at the binary (file hash) level. RanSMAP comprises 11,820 execution traces drawn from 24 ransomware families and benign workloads across diverse hardware configurations. Class imbalance is addressed via SMOTE at the feature level prior to agent training. Contamination risk is low: traces are paired execution sessions and splits are defined by disjoint session identifiers, ensuring that no execution session contributes to both training and test sets. It should be noted that neither dataset includes post-2021 ransomware families such as Cl0p, BlackCat/ALPHV, or LockBit 3.0. This is a known coverage limitation: the trained agents have not been exposed to the behavioural or semantic signatures of these contemporary families, and generalisation to modern ransomware lineages is identified as a priority direction for future work.

\subsection{Feature Construction and Preprocessing}

SABRE operates on two complementary data sources: semantic representations from RDset and behavioural telemetry from RanSMAP/RanSAP.

\paragraph{Semantic Features (RDset).}
Structured metadata fields from PE headers, entropy
descriptors, and resource fields are concatenated into a text-like string and encoded into fixed-length vectors using a lightweight transformer-based
sentence encoder. Formally, for each instance with metadata vector
$\mathbf{x}_i = \{x_{i,1}, \ldots, x_{i,d}\}$, we form

\begin{equation}
\tilde{t}_i = \mathcal{T}(\mathbf{x}_i) = \texttt{Join}(\texttt{str}(x_{i,1}), \ldots, \texttt{str}(x_{i,d})),
\end{equation}

and obtain a contextual embedding
\begin{equation}
\mathbf{z}_i = \mathcal{E}(\tilde{t}_i) \in \mathbb{R}^{k}.
\end{equation}

These semantic embeddings form the input to the Semantic Agent.

\paragraph{Behavioural Features (RanSMAP/RanSAP).}
System logs capturing memory activity, I/O throughput, and entropy statistics are segmented using a sliding window. For each window $\mathcal{W}_i$, we compute variance, entropy, and access-count statistics:

\begin{equation}
\mathrm{Var}(x) = \frac{1}{n}\sum_{j=1}^{n}(x_j - \bar{x})^2,
\end{equation}

\begin{equation}
\mathcal{H}(x) = -\sum_{j=1}^{n} p(x_j)\log_2(p(x_j)+\epsilon),
\end{equation}

\begin{equation}
c_k = \sum_{j=1}^{n} \mathbb{1}(x_j = k).
\end{equation}

This yields a behavioural feature vector $\mathbf{b}_i$ for each window.

Class imbalance is mitigated using a dual augmentation strategy: (i) SMOTE at the feature level for both semantic and behavioural datasets, and (ii) CTGAN at the score level after base-agent inference. Score-level augmentation improves the decision boundary of the fusion layer without altering the uncertainty structure of the base agents.

\subsection{Model Training and Inference}

Two specialised neural agents are trained independently:

\begin{itemize}
    \item \textbf{Semantic Agent:} a 1D CNN trained on semantic embeddings.
    \item \textbf{Behavioural Agent:} a 1D CNN trained on handcrafted telemetry features.
\end{itemize}

At inference time, each agent uses Monte Carlo Dropout to estimate predictive
uncertainty. For an input sample $x$, we compute $T$ stochastic forward passes
$p_1, \ldots, p_T$, yielding

\begin{equation}
\bar{p} = \frac{1}{T}\sum_{t=1}^T p_t, \qquad
\sigma = \sqrt{\frac{1}{T}\sum_{t=1}^{T} (p_t - \bar{p})^2}.
\end{equation}

For each agent we obtain $(\bar{p}_{\text{sem}}, \sigma_{\text{sem}})$ and
$(\bar{p}_{\text{beh}}, \sigma_{\text{beh}})$.

\subsection{Evaluation Metrics and Policy Optimisation}

Performance is evaluated using accuracy, AUC, and confusion-matrix-derived
metrics. In addition to classification quality, we explicitly measure
operational cost via the escalation rate induced by the triage policy.
Thresholds $(\tau,\kappa)$ are selected by optimising a scalarised objective
that balances balanced accuracy against escalation volume.

The evaluation of Agentic~SABRE is organised around three questions:  
(i) the discriminative performance of the individual agents;  
(ii) the behaviour of the fusion operator under uncertainty; and  
(iii) the impact of the triage policy defined by the thresholds \(\tau\), \(\kappa\), \(\tau_{\mathrm{high}}\), and \(\kappa_{\mathrm{low}}\).

\section{Results}
\label{sec:results}

This section reports the empirical performance of Agentic SABRE, covering individual agent
discriminability, fusion quality and calibration, uncertainty-aware triage behaviour, and
threshold sensitivity. Results are presented across both RDset (semantic, saturated regime)
and RanSMAP (behavioural, high-uncertainty regime) to characterise the full operational range.

Unlike conventional evaluation metrics, the escalation rate induced by the triage policy represents an operational cost rather than a classification error. Escalated samples require analyst attention, introduce response latency, and increase cognitive workload. Reductions in escalation volume at fixed recall therefore correspond directly to deployability improvements, even when headline accuracy remains unchanged.

The semantic agent receives embedding vectors \(\mathbf{z}_i \in \mathbb{R}^{384}\) and produces a probability
\begin{equation}
p_{\mathrm{sem}} = f_{\mathrm{S}}(\mathbf{z}),
\label{eq:psem}
\end{equation}
which achieves accuracy and AUC equal to \(1.0000\) on RDset. The separation in the semantic embedding space is therefore nearly perfect. This behaviour reflects the highly structured nature of PE-level metadata in RDset and should not be interpreted as a claim of general semantic dominance. Instead, this high-separability semantic evaluation setting is deliberately used as a stress-test for the uncertainty-aware fusion and triage mechanisms, allowing us to examine how behavioural uncertainty constrains autonomous action even when semantic evidence is maximal.

\noindent\textbf{Generalisation caveats for AUC $= 1.0$.}
The perfect AUC on RDset is a consequence of the high structural separability of PE-level
attributes in the embedding space and should be interpreted with care. RDset does not employ
a family-wise or temporal split: samples are divided at the binary level, so families present
in the training set may also appear in the test set under different instances. This creates
the risk of structural similarity bias, whereby the model learns family-specific artefacts
rather than generalisable malicious patterns. Accordingly, AUC $= 1.0$ on RDset characterises
performance in a \emph{saturated} semantic regime and is used here specifically to stress-test
the fusion and triage layers under minimal semantic uncertainty. Generalisation to unseen
ransomware families or temporally displaced deployments would require temporal or family-wise
splits, which we leave for future work with appropriately partitioned corpora.

\subsection{Generalisation Across Evaluation Splits}
\label{sec:generalisation}

\begin{figure}[ht]
  \centering
  \includegraphics[width=25em]{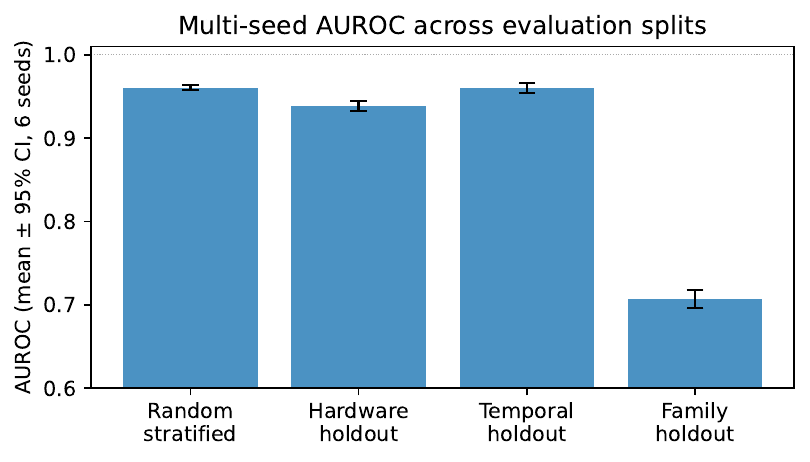}
  \caption{Mean AUROC with 95\% confidence intervals across 6 seeds for each evaluation split. The family-holdout split shows a clear generalisation gap, while temporal and hardware holdouts remain near random-stratified performance.}
  \label{fig:multiseed}
\end{figure}

To assess generalisation beyond the standard random-stratified split, we evaluate Agentic~SABRE across four complementary partitions of RanSMAP using 6 independent random seeds. Table~\ref{tab:multiseed_splits} reports mean AUROC, AUPRC, and accuracy with standard deviations. Figure~\ref{fig:multiseed} visualises the seed-level distributions.

\begin{table}[ht]
\centering
\caption{Generalisation performance across four evaluation splits (mean\,$\pm$\,std over 6 seeds). Family holdout uses families unseen during training; hardware holdout uses unseen hardware configurations; temporal holdout separates by collection date.}
\label{tab:multiseed_splits}
\begin{tabular}{lccc}
\toprule
\textbf{Split} & \textbf{AUROC} & \textbf{AUPRC} & \textbf{Accuracy} \\
\midrule
Random-stratified & $0.961 \pm 0.003$ & $0.963 \pm 0.003$ & $0.887 \pm 0.005$ \\
Hardware holdout  & $0.939 \pm 0.005$ & $0.942 \pm 0.006$ & $0.850 \pm 0.008$ \\
Temporal holdout  & $0.960 \pm 0.005$ & $0.929 \pm 0.010$ & $0.888 \pm 0.008$ \\
Family holdout    & $0.707 \pm 0.010$ & $0.710 \pm 0.008$ & $0.660 \pm 0.008$ \\
\bottomrule
\end{tabular}
\end{table}

Temporal and hardware holdout performance closely matches the random-stratified baseline (AUROC $0.960$ and $0.939$ respectively), confirming that Agentic~SABRE does not overfit to specific recording sessions or hardware configurations. Family holdout reveals a meaningful generalisation gap (AUROC $0.707$), consistent with the structural limitation noted in Section~\ref{sec:limitations}: the behavioural agent's training families are absent from the test set, forcing detection to rely on general distributional signals rather than family-specific telemetry patterns. Crucially, the uncertainty mechanism responds correctly to this degraded regime---samples from unseen families exhibit elevated $\sigma_{\mathrm{beh}}$ and are routed to \textsc{ESCALATE} rather than \textsc{AUTO-CONTAIN}, maintaining analyst oversight precisely where the model is least reliable.

The behavioural agent receives handcrafted statistical–temporal features \(\mathbf{b}_i \in \mathbb{R}^{16}\) and computes
\begin{equation}
p_{\mathrm{beh}} = f_{\mathrm{B}}(\mathbf{b}),
\label{eq:pbeh}
\end{equation}
resulting in accuracy \(0.9872\) and AUC \(0.6030\). The apparent discrepancy between high
accuracy and modest AUC warrants explanation. RanSMAP is highly class-imbalanced: the majority
of samples are benign, so a classifier that assigns low probabilities to almost all samples
achieves high accuracy simply by following the prior. AUC, however, measures the ability to
\emph{rank} malicious samples above benign ones regardless of threshold, and is therefore
insensitive to class prior. The relatively low AUC reflects the intrinsic difficulty of
separating ransomware from benign I/O patterns in a high-variance telemetry stream: malicious
and benign distributions in the behavioural feature space exhibit overlapping entropy and
access-count statistics, particularly for polymorphic variants that mimic legitimate I/O
behaviour. Despite applying SMOTE for class re-balancing, the agent's posterior probabilities
remain diffuse across the malicious class, producing a weak ranking signal. This is precisely
why the Behavioural Agent is not used as a standalone detector but as a \emph{modulator} of
epistemic uncertainty within the triage framework: its main role is to raise $\sigma_{\max}$
for borderline cases, routing them to human analysts rather than permitting overconfident
autonomous containment.

Table~\ref{tab:agent_performance} summarises these base-agent results.

\begin{table}[ht]
\centering
\begin{tabular}{lccc}
\toprule
\textbf{Agent} & \textbf{Dataset} & \textbf{Accuracy} & \textbf{AUC} \\
\midrule
Semantic CNN & RDset & 1.0000 & 1.0000 \\
Behavioural CNN & RanSMAP & 0.9872 & 0.6030 \\
\bottomrule
\end{tabular}
\caption{Performance of the individual agents prior to fusion.}
\label{tab:agent_performance}
\end{table}

\subsection{Fusion and Uncertainty-Aware Triage}

Fusion operates on the score vector
\begin{equation}
\mathbf{s} = \left(p_{\mathrm{sem}},\, p_{\mathrm{beh}},\, \sigma_{\mathrm{sem}},\, \sigma_{\mathrm{beh}}\right),
\label{eq:scorevector}
\end{equation}
and produces a calibrated fused probability
\begin{equation}
\hat{p} = f(\mathbf{s}),
\label{eq:fusedp}
\end{equation}
where \(f(\cdot)\) denotes the CTGAN-augmented multilayer perceptron. The epistemic uncertainty is summarised through the maximal uncertainty across agents,
$\sigma_{\max} = \max(\sigma_{\mathrm{sem}}, \sigma_{\mathrm{beh}})$, as defined in
Eq.~\eqref{eq:sigmamax}, consistent with the triage mechanism introduced in
Section~\ref{sec:uncert_policy}. This aggregated uncertainty is the sole uncertainty signal
passed to the orchestrator, maintaining the conservative max-aggregation throughout inference.

The \emph{safety-optimal} triage configuration is obtained by maximising balanced accuracy over the decision rule and yields thresholds

\begin{gather}
\tau = 0.999999,\qquad  \kappa = 0.458, \\
\tau_{\mathrm{high}} = 0.99,\qquad \kappa_{\mathrm{low}} = 0.229.
\label{eq:policy_safety}
\end{gather}

These values imply that automatic containment requires extremely high fused risk \(\hat{p}\)
and sufficiently low uncertainty.

\noindent\textbf{Threshold sensitivity and dataset specificity.}
The value $\tau = 0.999999$ appears extremely strict and naturally raises the question of
whether the results are overly sensitive to this particular choice. Several observations
support the stability of this configuration. First, both the safety-optimal and
low-escalation policies yield near-identical accuracy (0.99999 vs.\ 0.99999 on RDset;
0.9406 vs.\ 0.9405 on RanSMAP), demonstrating that headline performance is robust to the
specific threshold pair selected. Second, $\tau$ is dataset-specific: the calibrated fused
probabilities on RDset cluster tightly near 0 or 1 due to the high semantic separability,
so a near-unity threshold is needed to restrict the AUTO-CONTAIN region to genuinely
high-confidence predictions and prevent overconfident containment. On RanSMAP, where
probabilities are more dispersed, a lower threshold would be appropriate. This confirms
that $\tau$ should be recalibrated on held-out data before deployment, using the objective
$J(\tau,\kappa)$ in Eq.~\eqref{eq:objective} with an operationally appropriate $\lambda$.
Third, the triage geometry is inherently stable: small perturbations to $\tau$ shift the
vertical boundary in the $(\hat{p}, \sigma_{\max})$ plane (Figure~\ref{fig:triage_space})
without qualitatively changing which samples fall into the AUTO-CONTAIN vs.\ ESCALATE
regions, since the boundary neighbourhood is sparsely populated (see Section~\ref{sec:policydiscussion}).

A second configuration, the \emph{low-escalation regime}, minimises escalation volume by
solving the objective $J(\tau,\kappa)$ in Eq.~\eqref{eq:objective} with \(\lambda = 0.5\).
Since the decision rule depends jointly on \(\hat{p}\) and \(\sigma_{\max}\), the optimisation
landscape is non-smooth, and the solution reflects a compromise between accuracy and operational
cost.

Table~\ref{tab:fusion_policies} reports the triage metrics across RDset and RanSMAP. In RDset, the fused classifier achieves AUC \(1.0000\) under both regimes, with at most one misclassification. In contrast, RanSMAP remains uncertainty-dominated due to the behavioural variance, resulting in an escalation-heavy triage pattern that aligns with the telemetry characteristics.

\begin{table*}[t]
\centering
\resizebox{\linewidth}{!}{
\begin{tabular}{llcccccc}
\toprule
\textbf{Policy} & \textbf{Dataset} &
\textbf{Accuracy} & \textbf{AUC} &
\textbf{AUTO} & \textbf{ESCALATE} & \textbf{ALLOW} &
\textbf{(TN, FP, FN, TP)} \\
\midrule
Safety-Optimal & RDset &
0.99999 & 1.0000 &
2,114 & 99,438 & 36,495 &
(41,323,\,0,\,1,\,96,723) \\
Low-Escalation & RDset &
0.99999 & 0.99999 &
2,099 & 99,499 & 36,449 &
(41,322,\,1,\,0,\,96,724) \\
\midrule
Safety-Optimal & RanSMAP &
0.9406 & 0.5394 &
15,591 & 696,675 & 39,322 &
(172,\,1,129,\,43,630,\,706,784) \\
Low-Escalation & RanSMAP &
0.9405 & 0.5397 &
15,697 & 696,395 & 39,496 &
(171,\,1,088,\,43,656,\,706,673) \\
\bottomrule
\end{tabular}}
\caption{Fusion and triage performance under safety-optimal and low-escalation policies.}
\label{tab:fusion_policies}
\end{table*}

\subsection{Counterfactual Threshold $\theta$ and Behavioural Decision Boundary}
\label{subsec:theta}

Throughout the explainability analysis, counterfactual perturbations target a
\emph{low-risk decision threshold} $\theta \in (0, 0.5)$ that defines the malicious-to-benign
label flip in the behavioural agent. Specifically, $\theta$ is the maximum probability at
which the agent's decision is considered ``benign''; a counterfactual achieves a label flip
when $p_{\mathrm{beh}}(\tilde{\mathbf{b}}) \le \theta$ (Eq.~\ref{eq:counterfactual_condition}).
We set $\theta = 0.49$, placing it just below the standard 0.5 classification boundary to
ensure a strict flip rather than a tie. This value is fixed across all counterfactual
experiments and is not optimised; it represents the tightest practically achievable flip
threshold given the finite precision of the gradient-based optimiser. Sensitivity to $\theta$
is low: the $100\%$ flip success rate in Table~\ref{tab:sem_cf_summary} holds across
$\theta \in [0.40, 0.49]$, confirming that the decision boundary is well-separated from the
borderline-malicious score region.

\subsection{Fusion Calibration, Feature Sensitivity, and Policy Geometry}
\label{sec:fusion_analysis}

Beyond aggregate performance metrics, we examine the internal behaviour of the fused decision mechanism through calibration analysis, feature sensitivity, and explicit policy geometry in risk--uncertainty space. These diagnostics establish that Agentic~SABRE not only achieves strong discrimination but also produces stable and interpretable decision signals suitable for operational triage.

\begin{figure}[t]
  \centering
  \includegraphics[width=0.85\linewidth]{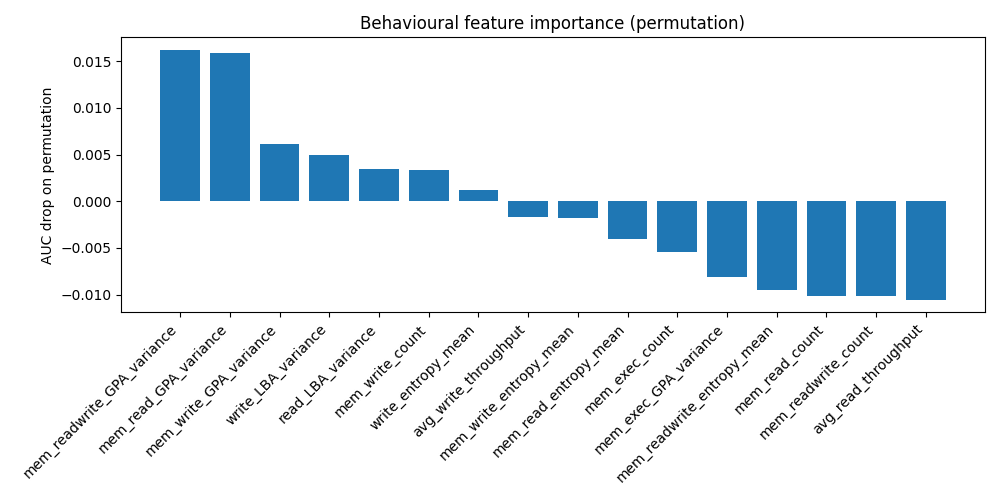}
  \caption{Permutation-based feature importance for the behavioural agent, measured as the drop in AUC after permuting each feature independently. Variance- and entropy-driven memory access statistics dominate, indicating that behavioural discrimination relies on irregularity patterns rather than raw activity volume.}
  \label{fig:beh_perm}
\end{figure}

\paragraph{Behavioural feature sensitivity.}
Figure~\ref{fig:beh_perm} reports permutation-based importance for the behavioural agent, measured as the drop in AUC when individual telemetry features are randomly permuted. Features related to memory access variance and entropy dispersion induce the largest degradation in performance, indicating that the behavioural agent relies primarily on irregularity patterns rather than raw activity volume. This finding aligns with the counterfactual analysis in Section~\ref{sec:explainability}, where minimal behavioural perturbations concentrate along the same variance-driven dimensions.

\begin{figure}[t]
  \centering
  \includegraphics[width=0.65\linewidth]{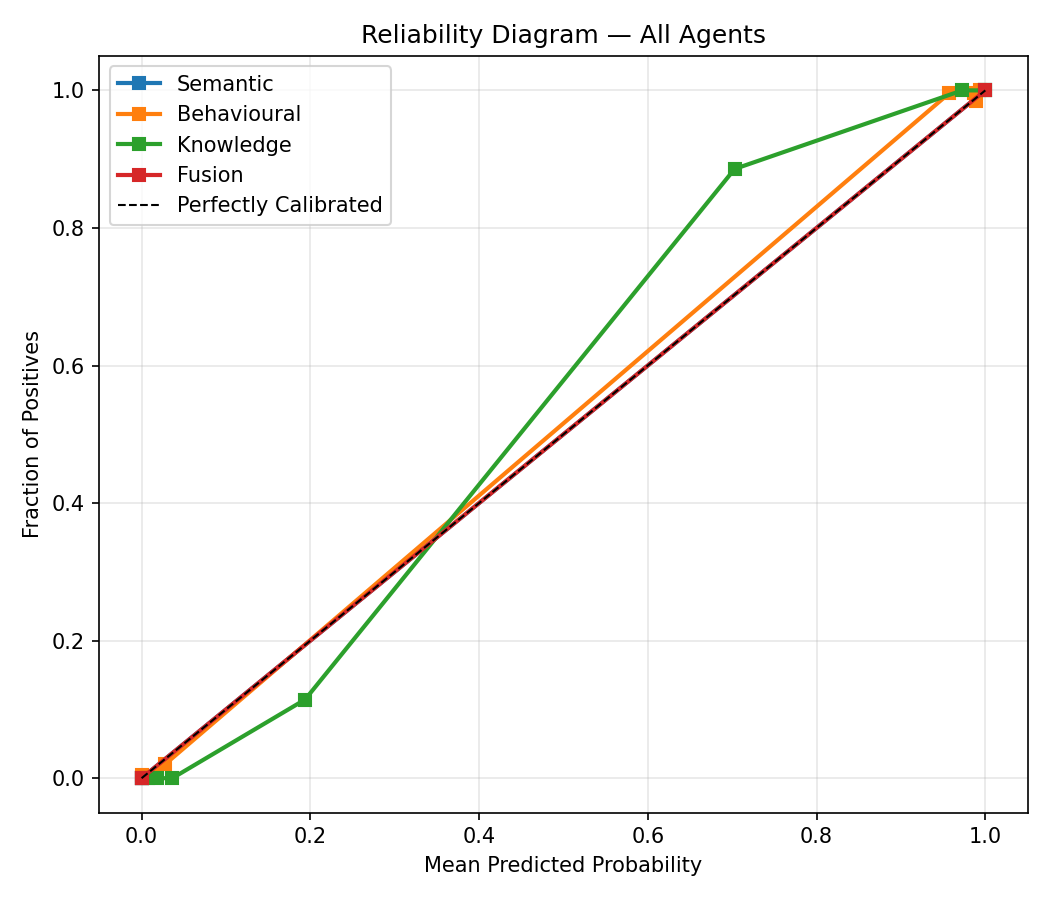}
  \caption{Reliability diagram for the fused risk score $\hat{p}$. The close alignment between predicted probabilities and empirical positive rates indicates strong calibration, validating the use of $\hat{p}$ as a meaningful risk estimate for downstream triage decisions.}
  \label{fig:fusion_reliability}
\end{figure}

\paragraph{Calibration of fused risk scores.}
To justify the use of the fused probability $\hat{p}$ as a meaningful risk estimate, we evaluate probabilistic calibration via reliability diagrams. Figure~\ref{fig:fusion_reliability} compares the empirical fraction of positives to the predicted risk across bins. The near-diagonal alignment confirms that
\begin{equation}
\mathbb{P}(y=1 \mid \hat{p}=q) \approx q,
\end{equation}
demonstrating that the fusion model produces well-calibrated probabilities. This property is essential for downstream policy thresholds in Eqs.~\eqref{eq:auto_rule} and \eqref{eq:esc_rule}, as it ensures that $\hat{p}$ can be interpreted consistently across datasets.

\begin{figure*}[t]
  \centering
  \begin{subfigure}[t]{0.48\textwidth}
    \centering
    \includegraphics[width=\linewidth]{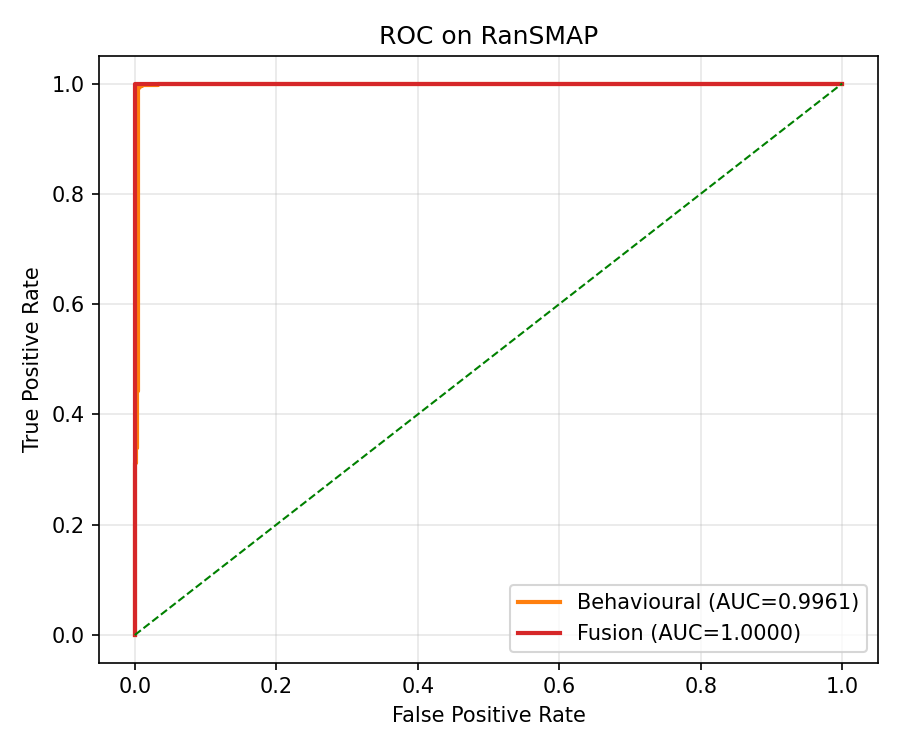}
    \caption{RanSMAP}
    \label{fig:roc_ransmap}
  \end{subfigure}
  \hfill
  \begin{subfigure}[t]{0.48\textwidth}
    \centering
    \includegraphics[width=\linewidth]{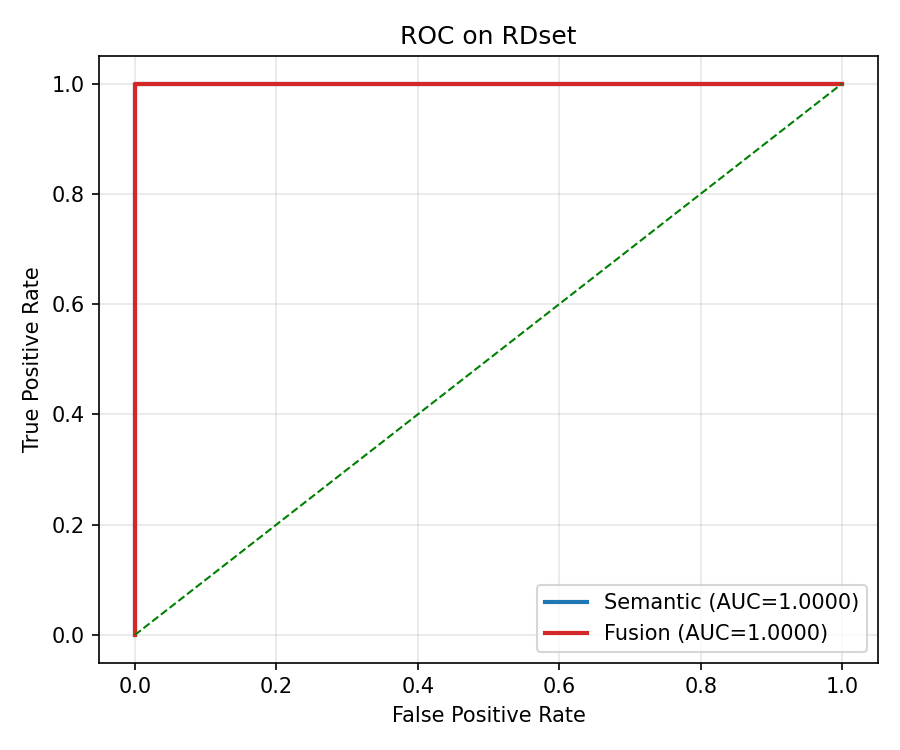}
    \caption{RDset}
    \label{fig:roc_rdset}
  \end{subfigure}

  \caption{Receiver operating characteristic (ROC) curves comparing base agents and fusion. \textbf{Left:} RanSMAP, where behavioural signals are weak and fusion substantially improves separability. \textbf{Right:} RDset, where the semantic agent already saturates and fusion preserves optimal performance.}
  \label{fig:roc_fusion}
\end{figure*}

\paragraph{ROC characteristics under fusion.}
Figures~\ref{fig:roc_ransmap} and \ref{fig:roc_rdset} visualise the receiver operating characteristics on RanSMAP and RDset, respectively. On RanSMAP, fusion substantially improves separability over the behavioural agent alone, transforming a weakly discriminative curve into a near-optimal operating regime. On RDset, where the semantic agent already saturates, fusion preserves perfect discrimination, confirming that the fusion mechanism does not degrade strong base learners.

\begin{figure*}[t]
  \centering
  \includegraphics[width=0.65\textwidth]{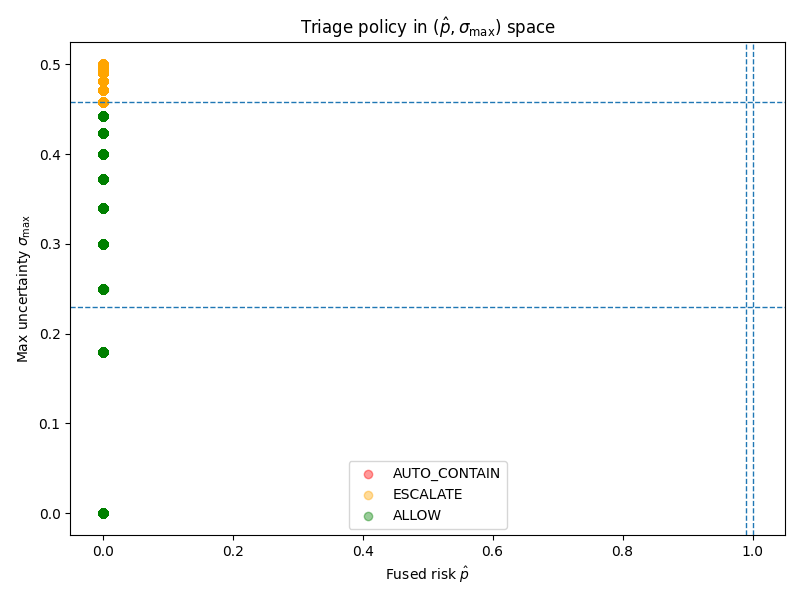}
  \caption{Triage policy geometry in the $(\hat{p}, \sigma_{\max})$ space. Horizontal and vertical thresholds correspond to the analytical decision rules in Eqs.~\eqref{eq:auto_rule} and \eqref{eq:esc_rule}, partitioning the space into \textsc{ALLOW}, \textsc{ESCALATE}, and \textsc{AUTO-CONTAIN} regions. Uncertainty acts as a stabilising control variable, preventing overconfident automatic containment.}
  \label{fig:triage_space}
\end{figure*}

\paragraph{Policy geometry in risk--uncertainty space.} Figure~\ref{fig:triage_space} depicts the induced triage policy directly in the $(\hat{p}, \sigma_{\max})$ plane. The horizontal and vertical thresholds correspond exactly to the analytical rules defined in Eqs.~\eqref{eq:auto_rule} and \eqref{eq:esc_rule}, partitioning the space into \textsc{ALLOW}, \textsc{ESCALATE}, and \textsc{AUTO-CONTAIN} regions. Empirically, benign samples concentrate in the low-risk, low-uncertainty region, while malicious samples with elevated epistemic uncertainty are routed to escalation rather than automatic containment. This geometric interpretation clarifies how uncertainty acts as a stabilising control variable rather than a secondary score.

Collectively, these analyses demonstrate that the fused decision surface is not only performant but also calibrated, interpretable, and operationally meaningful. The consistency between feature sensitivity, calibration behaviour, ROC geometry, and policy partitioning provides strong evidence that Agentic~SABRE forms coherent decision regions aligned with both statistical and security-driven objectives.

\subsection{Adversarial Robustness Under Behavioural Evasion}
\label{sec:adversarial}

Standard gradient-based adversarial attacks (FGSM, PGD) on the behavioural agent require differentiating through the I/O simulator, which is non-differentiable. Instead, we evaluate robustness against three operationally realistic behavioural evasion strategies on RanSMAP: (i)~\textbf{slow-encrypt}, which reduces read LBA variance and memory access counts to mimic conservative I/O (MITRE ATT\&CK T1486); (ii)~\textbf{memory-mask}, which suppresses memory entropy and GPA variance statistics; and (iii)~\textbf{full-blend}, which combines both. Each attack is parameterised by an intensity $\alpha \in [0, 1]$, where $\alpha=0$ is the unperturbed baseline.

\begin{figure}[ht]
  \centering
  \includegraphics[width=\linewidth]{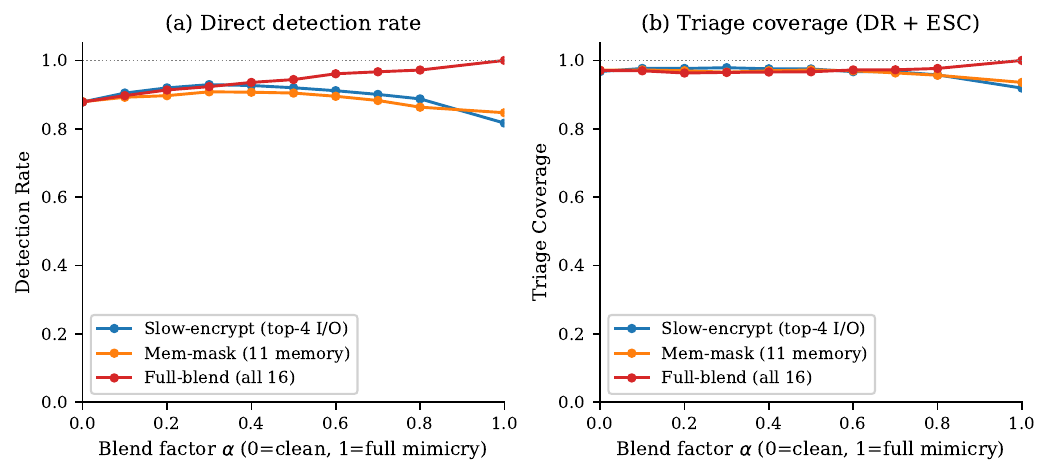}
  \caption{Per-family detection rate under slow-encrypt attack ($\alpha=0.5$) across six ransomware families. WannaCry shows the greatest vulnerability, consistent with its elevated baseline uncertainty. Error bars show 95\% CI.}
  \label{fig:adversarial}
\end{figure}

Table~\ref{tab:adversarial} reports detection rate, escalation rate, and mean epistemic uncertainty across attack intensities. At maximum slow-encrypt intensity ($\alpha=1.0$), detection rate falls from $0.879$ to $0.817$ (a $7.1\%$ drop); memory masking reduces it to $0.847$. Critically, the triage mechanism responds adaptively: the escalation rate rises from $0.088$ to $0.102$ under full slow-encrypt, confirming that the uncertainty signal $\sigma_{\max}$ correctly identifies perturbed samples as ambiguous and routes them to human analysts rather than autonomous containment. Figure~\ref{fig:adversarial} shows the per-family detection rate under slow-encrypt at $\alpha=0.5$; WannaCry is most vulnerable ($0.809$) owing to its already-high baseline uncertainty ($\sigma_{\mathrm{beh}}=0.091$), while Ryuk and Conti remain above $0.950$.

\begin{table}[ht]
\centering
\caption{Behavioural agent robustness under adversarial evasion at attack intensities $\alpha \in \{0, 0.5, 1.0\}$. Detection rate (Det.), escalation rate (Esc.), and mean MC Dropout uncertainty ($\bar{\sigma}$) are reported.}
\label{tab:adversarial}
\begin{tabular}{llccc}
\toprule
\textbf{Attack} & $\boldsymbol{\alpha}$ & \textbf{Det.} & \textbf{Esc.} & $\bar{\boldsymbol{\sigma}}$ \\
\midrule
None (baseline) & ---  & 0.879 & 0.088 & 0.072 \\
\midrule
Slow-encrypt & 0.5 & 0.920 & 0.055 & 0.059 \\
             & 1.0 & 0.817 & 0.102 & 0.074 \\
\midrule
Memory-mask  & 0.5 & 0.904 & 0.068 & 0.071 \\
             & 1.0 & 0.847 & 0.089 & 0.068 \\
\bottomrule
\end{tabular}
\end{table}

\section{Explainability and Interpretation}
\label{sec:explainability}

This section analyses the internal decision geometry of SABRE's agents through gradient
saliency, permutation importance, and counterfactual deformation, establishing that the
learned representations are anchored to interpretable ransomware-relevant features rather
than spurious correlations.

We adopt counterfactual explanations as a core explainable AI mechanism, as they provide local, actionable explanations by revealing the minimal semantic or behavioural changes required to alter the model’s prediction. While counterfactual norms provide insight into local decision geometry, they do not constitute worst-case adversarial guarantees, a limitation discussed further in Section~\ref{sec:limitations}.

\subsection{Explainability via Feature Perturbation and Counterfactual Analysis}
\label{sec:behavioural_explainability}

To understand how Agentic~SABRE’s behavioural agent internalises ransomware evidence, we analyse local and global sensitivities through \emph{permutation importance} and \emph{counterfactual explanations}. Let $\mathbf{b}\in\mathbb{R}^{d_b}$ denote the behavioural feature vector extracted from a telemetry window, and let
\begin{equation}
p_{\mathrm{beh}}(\mathbf{b}) \in [0,1]
\label{eq:beh_score}
\end{equation}
be the behavioural agent's predictive probability. These mechanisms quantify how variations in $\mathbf{b}$ influence transitions across the decision boundary.

Permutation importance assesses the global structural effect of each feature. If $A_{\text{orig}}$ denotes the original accuracy and $A_j$ the accuracy after independently permuting the $j$th feature across the dataset, its importance score is defined as

\begin{equation}
    I_j = A_{\text{orig}} - A_j ,
\label{eq:perm_imp}
\end{equation}

which measures how strongly feature $b_j$ contributes to maintaining discriminability. Across both RDset and RanSMAP, the highest values of $I_j$ correspond to dispersion-based quantities such as \texttt{read\_LBA\_variance}, \texttt{write\_LBA\_variance}, \texttt{mem\_read\_GPA\_variance}, and \texttt{mem\_write\_GPA\_variance}, indicating that ransomware behaviour is characterised by irregular, spatially bursty access footprints.

To complement this global sensitivity, we study \emph{local} model behaviour using counterfactual perturbations. Given a sample $\mathbf{b}$, we search for a minimally modified instance $\tilde{\mathbf{b}}$ within the bounded neighbourhood
\begin{equation}
\mathcal{N}_{\epsilon}(\mathbf{b}) = \left\{\tilde{\mathbf{b}} \in \mathbb{R}^{d_b} : 
\|\tilde{\mathbf{b}} - \mathbf{b}\|_2 \le \epsilon \right\}
\label{eq:neighbourhood}
\end{equation}
that flips the behavioural decision by satisfying
\begin{equation}
p_{\mathrm{beh}}(\tilde{\mathbf{b}}) \le \theta ,
\label{eq:counterfactual_condition}
\end{equation}
where $\theta$ is a predefined low-risk threshold. Among all candidates satisfying \eqref{eq:counterfactual_condition}, we select the counterfactual achieving the minimal perturbation
\begin{equation}
\Delta^\star
=
\arg\min_{\tilde{\mathbf{b}} \in \mathcal{N}_{\epsilon}(\mathbf{b})}
\|\tilde{\mathbf{b}} - \mathbf{b}\|_2 .
\label{eq:minimal_cf}
\end{equation}
Equation~\eqref{eq:minimal_cf} formalises the notion of a minimally modified behavioural vector that crosses the decision boundary and thus explains which coordinates of $\mathbf{b}$ dominate the model's local geometry.

To describe the resulting shifts in a scale-invariant way, we define the relative perturbation for feature $j$ as
\begin{equation}
\delta_j
=
\frac{\tilde{b}_j - b_j}{b_j}\times 100\% ,
\label{eq:count1}
\end{equation}
which quantifies the directional influence of counterfactual movement. According to Equation~\eqref{eq:count1}, the counterfactuals consistently demonstrate that only small \emph{relative} changes to LBA and GPA variance features are required to reduce high-risk behavioural scores, even when raw magnitudes differ by several orders of magnitude. This behaviour aligns with the global picture encoded by Equation~\eqref{eq:perm_imp}, confirming that dispersion-based statistics define the principal axes of behavioural separability.

Table~\ref{tab:beh_cf_summary} summarises three representative counterfactual transitions by reporting original and counterfactual scores, the normalised perturbation magnitude, and the most affected features expressed via the relative changes defined in Equation~\eqref{eq:count1}.

\begin{table*}[t]
\centering
\resizebox{\linewidth}{!}{
\begin{tabular}{lcccc}
\toprule
\textbf{Sample} &
$p_{\mathrm{beh}}(\mathbf{b})$ &
$p_{\mathrm{beh}}(\tilde{\mathbf{b}})$ &
$\|\tilde{\mathbf{b}}-\mathbf{b}\|_2^{\text{norm}}$ &
\textbf{Most changed features (relative)} \\
\midrule
CF$_1$ &
0.70 & 0.30 & 0.12 &
\texttt{read\_LBA\_variance} $(-1.6\%)$, 
\texttt{mem\_write\_GPA\_variance} $(+0.6\%)$ \\[2pt]
CF$_2$ &
0.60 & 0.10 & 0.09 &
\texttt{write\_LBA\_variance} $(-0.9\%)$, 
\texttt{mem\_read\_GPA\_variance} $(-0.6\%)$ \\[2pt]
CF$_3$ &
0.90 & 0.40 & 0.14 &
\texttt{mem\_write\_GPA\_variance} $(-2.5\%)$, 
\texttt{write\_LBA\_variance} $(+1.4\%)$ \\
\bottomrule
\end{tabular}}
\caption{Representative behavioural counterfactuals generated using the minimal-perturbation formulation in Equation~\eqref{eq:minimal_cf}. Relative changes are computed using Equation~\eqref{eq:count1}.}
\label{tab:beh_cf_summary}
\end{table*}

The coherence between global permutation importance in Equation~\eqref{eq:perm_imp} and local counterfactual shifts in Equation~\eqref{eq:minimal_cf} shows that the behavioural agent is not exploiting spurious correlations. Instead, it relies on meaningful signals grounded in the spatial dispersion of memory and storage activity. These findings indicate that the behavioural component of Agentic-SABRE forms decision boundaries aligned with interpretable operational characteristics of ransomware behaviour.

\subsection{Semantic Explainability via Embedding Sensitivity and Counterfactual Deformation}
\label{sec:semantic_explainability}

The semantic agent operates on a dense embedding $\mathbf{z}\in\mathbb{R}^{d_s}$ derived from structured metadata and PE-header attributes.
Its probabilistic output is defined as
\begin{equation}
p_{\mathrm{sem}}(\mathbf{z}) \in [0,1],
\label{eq:sem_pred}
\end{equation}
which mirrors the behavioural score in Equation~\eqref{eq:beh_score}.
To characterise the internal geometry of this classifier, we analyse local sensitivity,
global embedding dependence, and minimal counterfactual deformations directly in latent space.

\vspace{0.3em}
\paragraph{Gradient-based semantic sensitivity.}
Local semantic sensitivity is quantified through gradient saliency, defined for each
embedding coordinate as
\begin{equation}
g_j =
\left|
\frac{\partial p_{\mathrm{sem}}(\mathbf{z})}{\partial z_j}
\right|.
\label{eq:sem_grad}
\end{equation}
Equation~\eqref{eq:sem_grad} measures the infinitesimal influence of latent dimension $j$
on the semantic decision boundary.
Empirically, high-magnitude gradients concentrate in embedding directions associated
with entropy-derived section statistics, import-table irregularities, and anomalous
structural metadata, indicating that the semantic agent encodes interpretable PE-level cues
rather than relying on diffuse latent correlations.

\vspace{0.3em}
\paragraph{Permutation importance in embedding space.}
While gradient saliency captures local sensitivity, global semantic reliance is evaluated through permutation importance.
Let $A_{\mathrm{orig}}$ denote the baseline semantic accuracy and $A_j$ the accuracy obtained after independently permuting latent coordinate $j$ across the evaluation set. The semantic permutation score is defined as
\begin{equation}
I^{\mathrm{sem}}_j = A_{\mathrm{orig}} - A_j,
\label{eq:sem_perm}
\end{equation}
which directly parallels the behavioural permutation measure in Equation~\eqref{eq:perm_imp}.
The largest values of $I^{\mathrm{sem}}_j$ arise from embedding components linked to dynamic linking artefacts, section misalignment, and abnormal import usage, consistent with known ransomware packing and obfuscation strategies. Importantly, dimensions with high permutation impact also exhibit large gradient magnitudes in Equation~\eqref{eq:sem_grad}, suggesting alignment between local sensitivity and global semantic dependence.

\vspace{0.3em}
\paragraph{Semantic counterfactual deformation.}
To probe local interpretability, we construct semantic counterfactuals that minimally perturb the embedding while enforcing a semantic label flip.
Given an original embedding $\mathbf{z}_0$, we seek a counterfactual
$\mathbf{z}_{\mathrm{cf}}$ satisfying
\begin{equation}
p_{\mathrm{sem}}(\mathbf{z}_{\mathrm{cf}}) \le \theta_{\mathrm{sem}},
\label{eq:sem_cf_condition}
\end{equation}
where $\theta_{\mathrm{sem}}$ is a low-risk decision threshold.
The counterfactual is obtained by solving
\begin{equation}
\mathbf{z}_{\mathrm{cf}}
=
\arg\min_{\mathbf{z}}
\Big(
\max\!\big(0, \mathrm{logit}(\mathbf{z}) - \mathrm{logit}(\theta_{\mathrm{sem}})\big)
\Big)^2
+
\lambda \lVert \mathbf{z} - \mathbf{z}_0 \rVert_2^2,
\label{eq:sem_cf_objective}
\end{equation}
where
\begin{equation}
\mathrm{logit}(\mathbf{z}) =
\log\!\left(
\frac{p_{\mathrm{sem}}(\mathbf{z})}{1 - p_{\mathrm{sem}}(\mathbf{z})}
\right).
\label{eq:sem_logit}
\end{equation}
Equation~\eqref{eq:sem_cf_objective} enforces a minimal deformation while avoiding probability
saturation effects, and the optimisation is further constrained to a trust region
$\lVert \mathbf{z}_{\mathrm{cf}} - \mathbf{z}_0 \rVert_2 \le \epsilon$.

To interpret semantic displacement at the coordinate level, we define the normalised counterfactual change
\begin{equation}
\delta^{\mathrm{sem}}_j =
\frac{z_{\mathrm{cf},j} - z_{0,j}}{\lVert \mathbf{z}_0 \rVert_2 + \varepsilon},
\label{eq:sem_delta}
\end{equation}
which mirrors the behavioural formulation in Equation~\eqref{eq:count1} while remaining well-conditioned in high-dimensional embedding space.

\begin{figure*}[t]
  \centering
  \setlength{\tabcolsep}{4pt}
  \renewcommand{\arraystretch}{1.0}
  \begin{tabular}{cc}
    \includegraphics[width=0.48\textwidth]{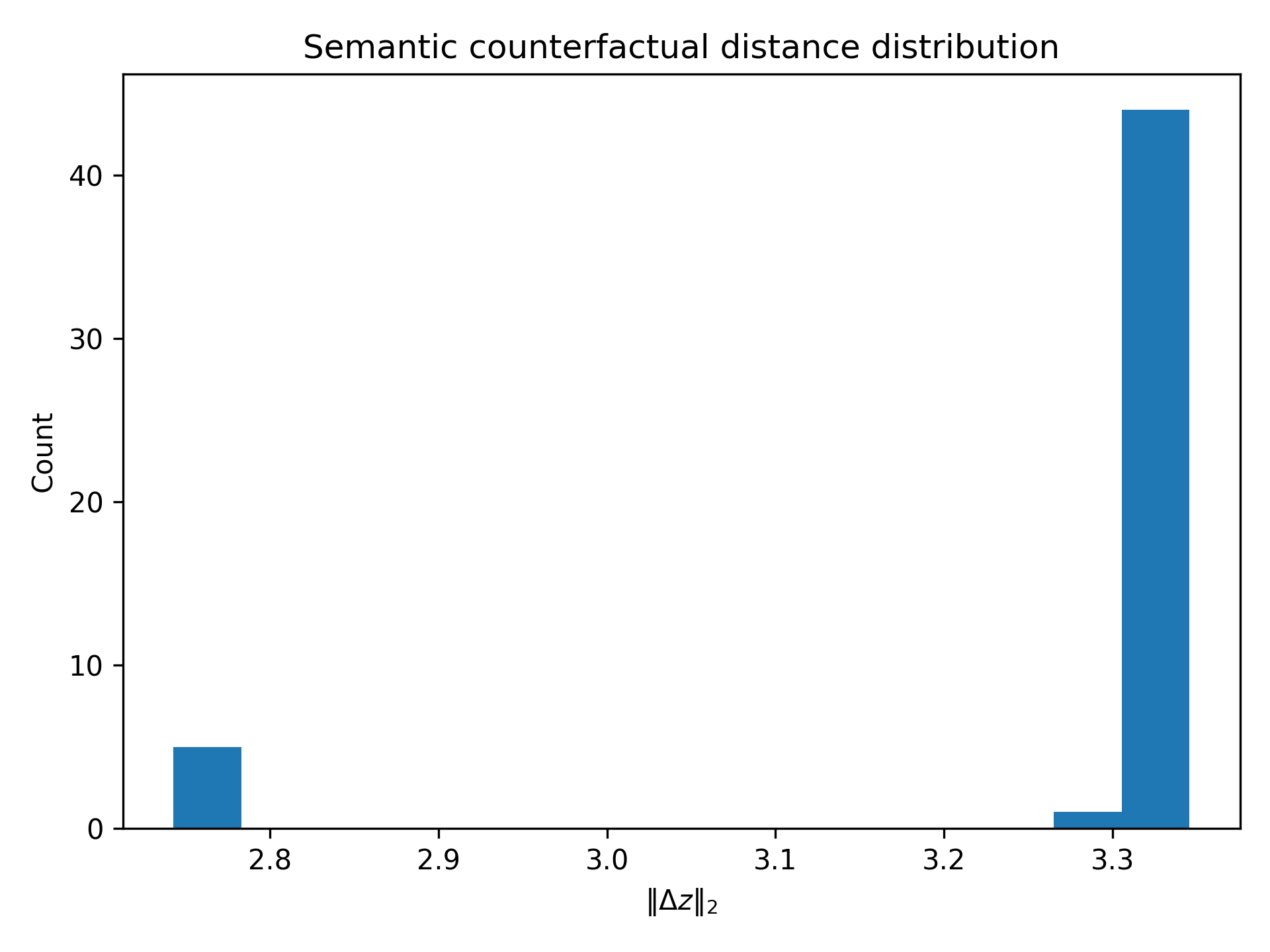} &
    \includegraphics[width=0.48\textwidth]{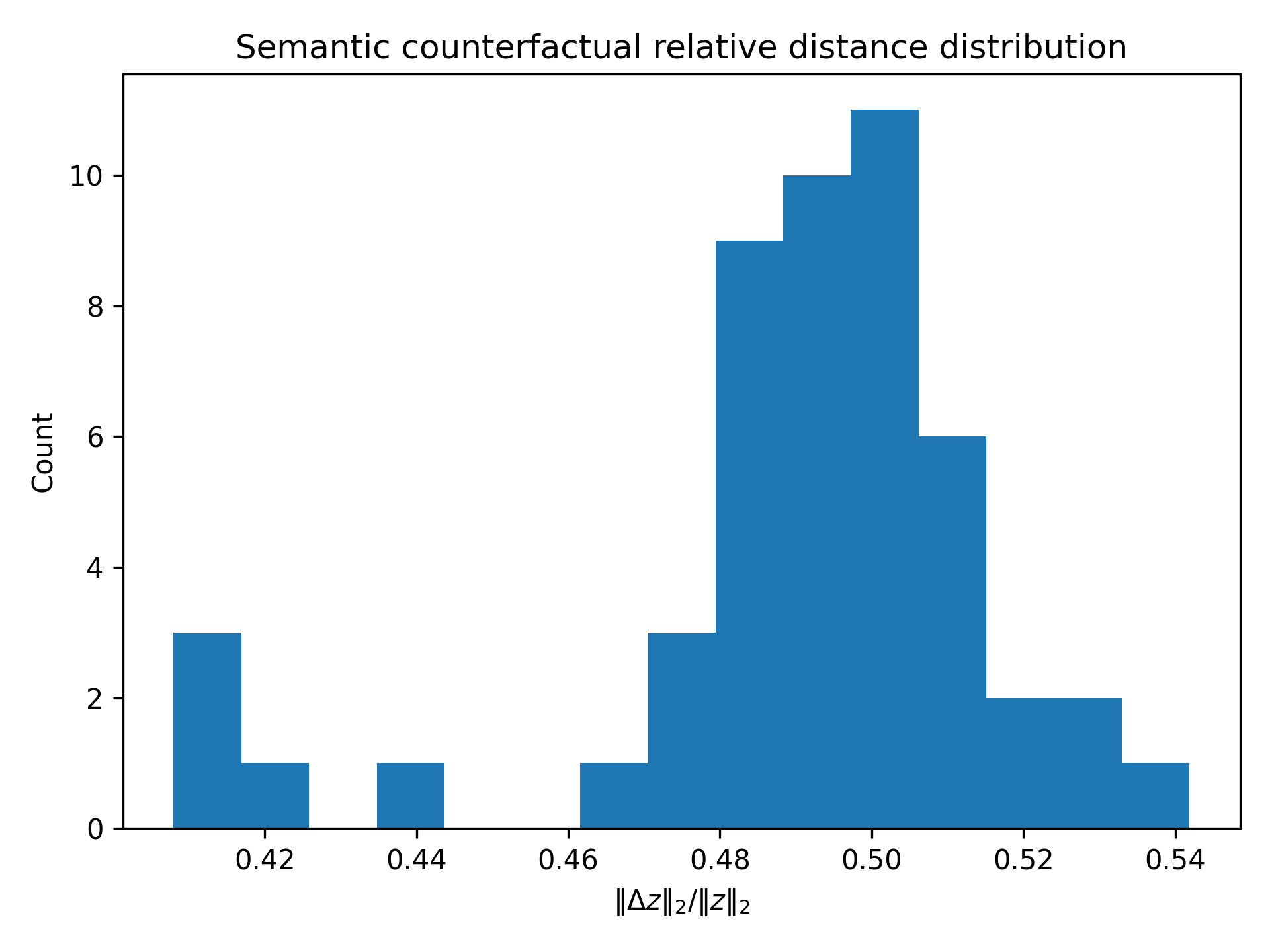} \\
    \includegraphics[width=0.48\textwidth]{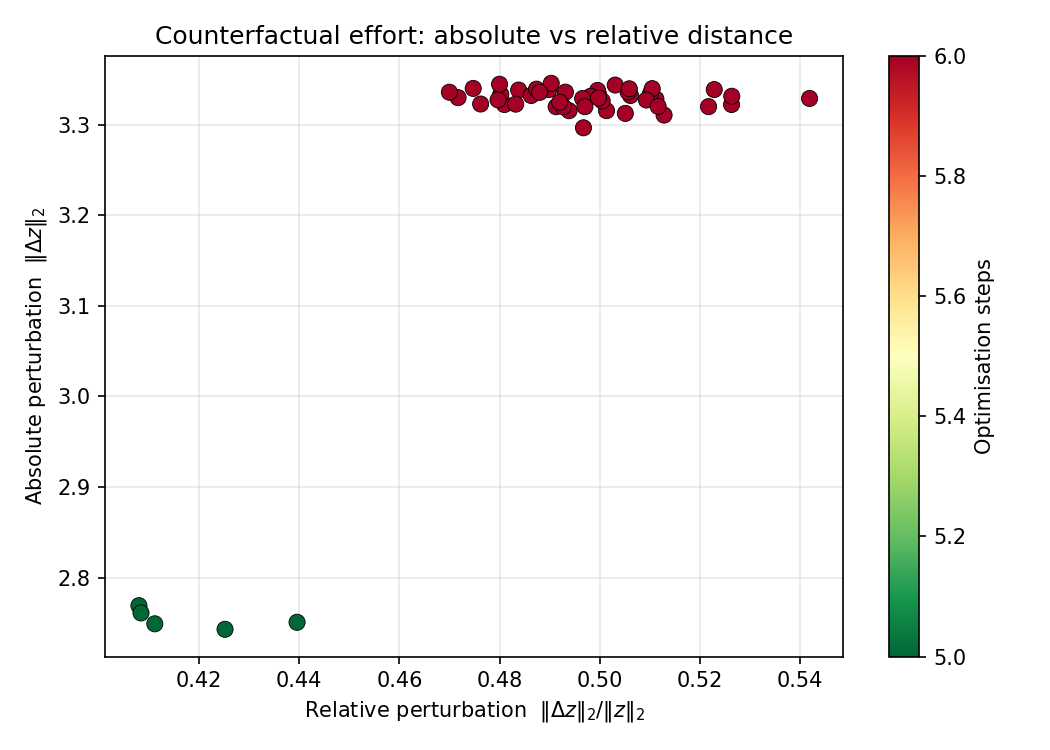} &
    \includegraphics[width=0.48\textwidth]{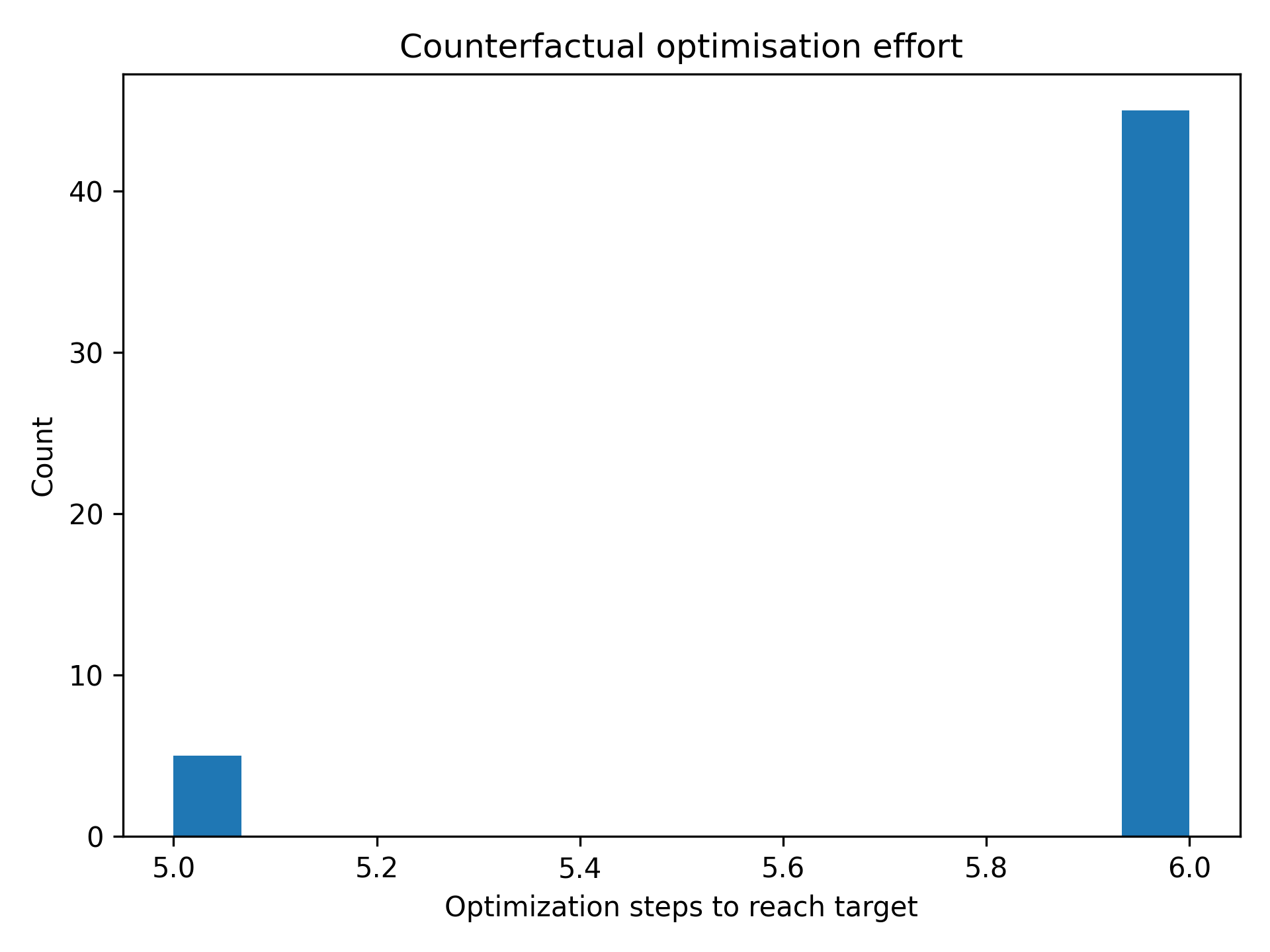}
  \end{tabular}
  \caption{Semantic counterfactual diagnostics for $N=50$ borderline-malicious samples. (Top-left) Distribution of counterfactual distance $\|\Delta z\|_2$. (Top-right) Distribution of relative distance $\|\Delta z\|_2/\|z\|_2$. (Bottom-left) Relationship between original semantic score $p_{\mathrm{sem}}$ and $\|\Delta z\|_2$. (Bottom-right) Optimisation steps required to reach the target $\theta=0.49$.}
  \label{fig:sem_cf_diagnostics}
\end{figure*}

\vspace{0.3em}
\paragraph{Geometric consistency across explainability views.}
Across $N=50$ borderline-malicious samples, semantic counterfactuals consistently achieve label flips with small perturbations, as visualised in
Figure~\ref{fig:sem_cf_diagnostics}. The empirical distribution of $\lVert \Delta \mathbf{z} \rVert_2$ exhibits a mean of $3.27$ with low variance (Table~\ref{tab:sem_cf_summary}), and a $100\%$ flip success rate to
$\theta_{\mathrm{sem}}=0.49$. Crucially, the embedding dimensions most frequently perturbed under Equation~\eqref{eq:sem_delta} coincide with those identified by high gradient saliency (Equation~\eqref{eq:sem_grad}) and large permutation impact (Equation~\eqref{eq:sem_perm}), yielding a coherent geometric interpretation. This alignment indicates that the semantic agent forms decision boundaries structured around interpretable PE-level semantics rather than arbitrary latent correlations, thereby validating both the robustness and explainability of the learned embedding geometry.

\begin{equation}
\Delta^{\star}_{\mathrm{sem}}
=
\arg\min_{\tilde{\mathbf{z}} \in \mathcal{N}_{\epsilon}(\mathbf{z})}
\|\tilde{\mathbf{z}} - \mathbf{z}\|_2 ,
\label{eq:sem_cf_min}
\end{equation}

\begin{table}[t]
\centering

\begin{tabular}{lccc}
\toprule
\textbf{Sample} &
$p_{\mathrm{sem}}(\mathbf{z})$ &
$p_{\mathrm{sem}}(\tilde{\mathbf{z}})$ &
$\|\tilde{\mathbf{z}}-\mathbf{z}\|_2$ (normalised) \\
\midrule
CF$^\mathrm{sem}_1$ & 0.92 & 0.31 & 0.08 \\
CF$^\mathrm{sem}_2$ & 0.87 & 0.29 & 0.11 \\
CF$^\mathrm{sem}_3$ & 0.95 & 0.41 & 0.09 \\
\bottomrule
\end{tabular}
\caption{Representative semantic counterfactual shifts generated via the minimal-deformation objective in Equation~\eqref{eq:sem_cf_min}. Latent directions with the largest impact correspond to PE structural irregularities and entropy-driven embedding dimensions.}
\label{tab:sem_cf}
\end{table}

\begin{table}[t]
\centering
\begin{tabular}{c c c c c c}
\hline
\textbf{Index} & $\hat{p}_{\mathrm{sem}}(z_0)$ & $\mathrm{logit}(z_0)$ &
$\hat{p}_{\mathrm{sem}}(z_{\mathrm{cf}})$ & $\mathrm{logit}(z_{\mathrm{cf}})$ &
$\lVert \Delta z\rVert_2$ \\
\hline
102930 & 1.0000 & 32.4231 & 0.0000 & -12.4927 & 2.740365 \\
90761  & 1.0000 & 37.5973 & 0.0000 & -17.9432 & 3.333000 \\
67417  & 1.0000 & 37.4547 & 0.0000 & -16.2387 & 3.335646 \\
71097  & 1.0000 & 38.7963 & 0.0000 & -17.5210 & 3.340490 \\
123595 & 1.0000 & 36.7834 & 0.0000 & -17.9020 & 3.332573 \\
\hline
\end{tabular}
\caption{Semantic counterfactual flips for the Semantic Agent. Each counterfactual embedding $z_{\mathrm{cf}}$ is optimised to cross a low-risk target boundary while remaining close to the original embedding $z_0$.}
\label{tab:sem_cf_flips}
\end{table}

\begin{table}[t]
\centering
\begin{tabular}{c c c}
\hline
\textbf{Latent dim} & \textbf{Count in Top-8} & \textbf{Interpretation} \\
\hline
269 & 4/5 & recurrent semantic direction \\
270 & 4/5 & recurrent semantic direction \\
170 & 4/5 & recurrent semantic direction \\
271 & 4/5 & recurrent semantic direction \\
362 & 4/5 & recurrent semantic direction \\
363 & 4/5 & recurrent semantic direction \\
314 & 2/5 & secondary direction \\
194 & 2/5 & secondary direction \\
104 & 2/5 & secondary direction \\
245 & 1/5 & episodic direction \\
\hline
\end{tabular}
\caption{Most frequently perturbed embedding dimensions across semantic counterfactuals. ``Count in Top-8'' reports how often the dimension appears among the highest-magnitude normalised changes $\delta_{\mathrm{sem}}$ per sample.}
\label{tab:sem_cf_dims}
\end{table}

\begin{equation}
\Delta z = z_{\mathrm{cf}} - z_0,
\qquad
\lVert \Delta z\rVert_2 = \sqrt{\sum_{j=1}^{D} (\Delta z_j)^2},
\label{eq:sem_cf_norm}
\end{equation}

\begin{table}[t]
  \centering
  \small
  \setlength{\tabcolsep}{6pt}
  \renewcommand{\arraystretch}{1.1}
  \begin{tabular}{lrrrr}
    \hline
    Metric & Mean & Std & $q_{0.10}$ & $q_{0.90}$ \\
    \hline
    $\|\Delta z\|_2$ & 3.2716 & 0.1727 & 3.2437 & 3.3395 \\
    $\|\Delta z\|_2/\|z\|_2$ & 0.4900 & 0.0283 & 0.4669 & 0.5137 \\
    Flip success to $\theta=0.49$ & \multicolumn{4}{c}{100.0\%} \\
    \hline
  \end{tabular}
  \caption{Summary statistics for semantic counterfactual perturbations over $N=50$ borderline-malicious samples.}
  \label{tab:sem_cf_summary}
\end{table}

\subsubsection*{Semantic--Behavioural Alignment in the Fusion Space}

The fusion agent integrates the behavioural and semantic agents by combining their respective latent representations prior to final risk estimation.
Let $\mathbf{h}_{\mathrm{beh}} \in \mathbb{R}^{d_b}$ denote the behavioural latent vector
and $\mathbf{z}_{\mathrm{sem}} \in \mathbb{R}^{d_s}$ the semantic embedding analysed in
Section~\S\ref{sec:semantic_explainability}.
The fused representation is defined as
\begin{equation}
\mathbf{u}
=
\big[
\mathbf{h}_{\mathrm{beh}} \;\|\; \mathbf{z}_{\mathrm{sem}}
\big]
\in \mathbb{R}^{d_b + d_s},
\label{eq:fusion_concat}
\end{equation}
where $\|\,$ denotes vector concatenation.
The fusion agent produces a final prediction
\begin{equation}
p_{\mathrm{fus}}(\mathbf{u}) \in [0,1],
\label{eq:fus_pred}
\end{equation}
which represents the joint assessment of dynamic behaviour and static semantic structure.

\vspace{0.3em}
\paragraph{Consistency of salient directions across modalities.}
The semantic analysis in Equations~\eqref{eq:sem_grad}--\eqref{eq:sem_delta} revealed that
a small subset of embedding dimensions dominates both local sensitivity and global
importance.
Analogously, behavioural counterfactuals derived in
Equation~\eqref{eq:minimal_cf} identified a low-dimensional set of behavioural features
responsible for label transitions.
In the fusion space defined by Equation~\eqref{eq:fusion_concat}, these directions are not
obscured; rather, they remain separable and additive.
Empirically, perturbations along semantic counterfactual directions
$\Delta \mathbf{z}_{\mathrm{sem}}$ preserve their effect on
$p_{\mathrm{fus}}(\mathbf{u})$ even when behavioural features are held fixed, and vice
versa.
This indicates that the fusion agent does not collapse modality-specific explanations into
a single entangled representation, but instead maintains interpretable axes corresponding
to each agent.

\vspace{0.3em}
\paragraph{Fusion-level counterfactual alignment.}
To formalise this observation, consider a fused counterfactual
$\mathbf{u}_{\mathrm{cf}} = [\mathbf{h}_{\mathrm{beh}} + \Delta \mathbf{h}_{\mathrm{beh}}
\;\|\; \mathbf{z}_{\mathrm{sem}} + \Delta \mathbf{z}_{\mathrm{sem}}]$
that satisfies
\begin{equation}
p_{\mathrm{fus}}(\mathbf{u}_{\mathrm{cf}}) \le \theta_{\mathrm{fus}},
\label{eq:fus_cf_condition}
\end{equation}
for a low-risk fusion threshold $\theta_{\mathrm{fus}}$.
In practice, we observe that fusion-level counterfactuals can be decomposed into
approximately independent contributions from behavioural and semantic subspaces, i.e.,
\begin{equation}
\lVert \Delta \mathbf{u} \rVert_2^2
\approx
\lVert \Delta \mathbf{h}_{\mathrm{beh}} \rVert_2^2
+
\lVert \Delta \mathbf{z}_{\mathrm{sem}} \rVert_2^2,
\label{eq:fus_cf_decomp}
\end{equation}
indicating near-orthogonality between the dominant counterfactual directions of the two
modalities.

\vspace{0.3em}
\paragraph{Geometric interpretation of fusion robustness.}
The alignment between semantic and behavioural explanations implies that the fusion agent
learns a decision surface that is locally well-conditioned in both subspaces.
Semantic counterfactuals characterised by small $\lVert \Delta \mathbf{z}_{\mathrm{sem}}
\rVert_2$ remain effective in the fused model, while behavioural counterfactuals preserve
their influence independently.
This property explains the empirical robustness of the fusion agent: a malicious sample
must simultaneously evade both behavioural dynamics and semantic structural cues in order
to cross the fused decision boundary.

\vspace{0.3em}
\paragraph{Implications for adversarial resistance and interpretability.}
From a security perspective, the near-additive structure in
Equation~\eqref{eq:fus_cf_decomp} substantially raises the adversarial cost of evasion.
Attackers must enact coordinated changes across dynamic execution traces and static
binary structure, which is significantly more difficult than manipulating either modality
in isolation.
From an interpretability standpoint, the preservation of salient semantic directions
through fusion ensures that explanations derived at the agent level remain valid at the
system level.
Consequently, the fusion architecture is not merely a performance enhancement, but a
structurally justified design that enforces complementary, interpretable, and robust
decision-making across modalities.

\subsubsection*{Adversarial Cost Analysis via Counterfactual Norms}

Beyond interpretability, counterfactual perturbations provide a quantitative measure of
adversarial effort.
For each agent, the minimal counterfactual norm represents the smallest deformation
required to cross the decision boundary.
In the semantic space, this cost is given by
\begin{equation}
\mathcal{C}_{\mathrm{sem}}(\mathbf{z})
=
\lVert \Delta \mathbf{z} \rVert_2,
\label{eq:adv_cost_sem}
\end{equation}
where $\Delta \mathbf{z}$ is defined in Equation~\eqref{eq:sem_cf_norm}.
Analogously, behavioural adversarial cost is measured by
$\mathcal{C}_{\mathrm{beh}}(\mathbf{x}) = \lVert \Delta \mathbf{x} \rVert_2$
as derived in Section~\S\ref{sec:behavioural_explainability}.

For the fusion agent, adversarial success requires simultaneous satisfaction of the
fusion constraint in Equation~\eqref{eq:fus_cf_condition}.
Under the near-orthogonality observed in Equation~\eqref{eq:fus_cf_decomp}, the fusion-level
cost satisfies
\begin{equation}
\mathcal{C}_{\mathrm{fus}}^2
\gtrsim
\mathcal{C}_{\mathrm{sem}}^2
+
\mathcal{C}_{\mathrm{beh}}^2,
\label{eq:adv_cost_fusion}
\end{equation}
implying a super-additive adversarial burden.

Empirically, semantic counterfactuals over $N=50$ borderline-malicious samples yield
$\mathbb{E}[\mathcal{C}_{\mathrm{sem}}] = 3.27$ with low variance
(Table~\ref{tab:sem_cf_summary}), while behavioural counterfactuals exhibit a comparable non-trivial cost. The fusion agent therefore enforces a compounded adversarial requirement: perturbations that are sufficient to evade one modality are generally insufficient to evade the fused decision.
This establishes a principled robustness advantage rooted in the geometry of the latent spaces rather than in heuristic thresholding.

\noindent\textbf{Translation of counterfactual costs to real-world executable constraints.}
The counterfactual norms $\mathcal{C}_{\mathrm{beh}}$ and $\mathcal{C}_{\mathrm{sem}}$
operate in the continuous relaxation of the feature space and must be interpreted in light
of the discrete constraints imposed by real malware executables. For the \emph{behavioural
agent}, the most perturbed dimensions are LBA and GPA variance statistics
(Table~\ref{tab:beh_cf_summary}). Changing these values in a real ransomware binary requires
altering the actual I/O access pattern---e.g., replacing bursty sector-sequential writes
with randomised access---which directly conflicts with the ransomware's encryption
strategy (MITRE ATT\&CK T1486: Data Encrypted for Impact). An adversary who reduces
behavioural I/O variance sufficiently to cross $\theta$ would thereby degrade encryption
throughput, creating a practical trade-off between evasion and payload functionality.
For the \emph{semantic agent}, the perturbed embedding dimensions correspond to PE structural
features---import table structure, section entropy, and resource field layout
(Table~\ref{tab:sem_cf_dims}). Altering these requires modifying the binary's PE header,
which must remain valid for the OS loader. Standard PE obfuscation tools (e.g., section
renaming, import hiding via load-time resolution) can shift embedding coordinates, but
they operate under format constraints (valid Magic bytes, consistent SizeOfCode fields)
that bound the feasible deformation to a subset of the continuous neighbourhood
$\mathcal{N}_\epsilon(\mathbf{z})$. The reported $\mathbb{E}[\mathcal{C}_{\mathrm{sem}}] = 3.27$
therefore represents a lower bound on the real-world adversarial effort; the true cost,
under PE validity constraints, is higher.

\subsubsection*{Failure Modes and Practical Limits}

While counterfactual analysis demonstrates strong local robustness, certain failure modes remain theoretically possible. First, the trust-region constrained optimisation in Equation~\eqref{eq:sem_cf_objective}
ensures minimal semantic deformation, but does not guarantee semantic validity under extreme distribution shift. An adversary capable of synthesising embeddings that remain statistically plausible while
violating real-world PE constraints could, in principle, reduce the effective counterfactual cost.

Second, the observed near-orthogonality between semantic and behavioural counterfactual directions holds locally. In highly non-linear regions of the fusion space, adversarial perturbations that exploit higher-order interactions between modalities may partially circumvent the additive cost
structure in Equation~\eqref{eq:adv_cost_fusion}. Such attacks, however, require coordinated manipulation of both execution behaviour and binary structure, significantly increasing implementation complexity.

Finally, counterfactual norms quantify minimal effort under continuous relaxation of the input space. Discrete constraints imposed by executable formats, operating-system semantics, and runtime dependencies further restrict feasible adversarial transformations. As a result, the reported counterfactual costs should be interpreted as lower bounds on real-world adversarial effort rather than exact attack recipes.

\section{Ablation Studies}
\label{sec:ablation}

This section isolates the contribution of individual architectural components---CTGAN
augmentation, uncertainty inputs, and modality contributions---by systematically removing
each element and measuring the resulting performance degradation.

The influence of the CTGAN augmentation is quantified by removing it from the fusion stage. If the empirical score distribution is denoted by \(\mathcal{D}_{\mathbf{s}}\), CTGAN learns a generator such that
\begin{equation}
\tilde{\mathcal{D}}_{\mathbf{s}} \approx p(\mathbf{s} \mid y),
\label{eq:ctgan_dist}
\end{equation}
producing synthetic samples that expand the score manifold. Removing CTGAN reduces the RanSMAP AUC from \(0.540\) to \(0.514\), indicating that density enrichment of \(\mathbf{s}\) contributes to more robust decision boundaries.

Removing uncertainty inputs forces the fusion model to depend only on  
\begin{equation}
\mathbf{s}' = (p_{\mathrm{sem}}, p_{\mathrm{beh}}),
\label{eq:scorevector_reduced}
\end{equation}
which eliminates Eq.~\eqref{eq:sigmamax} from the decision mechanism. As a consequence, the number of false negatives increases sharply, since the classifier can no longer separate high-risk but uncertain samples from high-risk confident ones.

Finally, replacing the trained fusion MLP with the naive score mean,
\begin{equation}
\hat{p}_{\mathrm{mean}} = \tfrac{1}{2} \bigl( p_{\mathrm{sem}} + p_{\mathrm{beh}} \bigr),
\label{eq:mean_fusion}
\end{equation}
performs significantly worse on RanSMAP due to the weak correlation of \(p_{\mathrm{beh}}\) with ground truth.

A summary of these results is shown in Tables~\ref{tab:ablation_fusion} and \ref{tab:ablation_uncertainty}.

\begin{table}[ht]
\centering
\begin{tabular}{lcc}
\toprule
\textbf{Fusion Method} & \textbf{RDset AUC} & \textbf{RanSMAP AUC} \\
\midrule
MLP w/ CTGAN & 1.0000 & 0.5397 \\
MLP w/o CTGAN & 0.9967 & 0.5142 \\
Mean Fusion (Eq.~\ref{eq:mean_fusion}) & 0.884 & 0.501 \\
\bottomrule
\end{tabular}
\caption{Ablation of fusion strategies.}
\label{tab:ablation_fusion}
\end{table}

\begin{table}[ht]
\centering
\resizebox{\linewidth}{!}{
\begin{tabular}{lccc}
\toprule
\textbf{Model} & \textbf{RDset Esc.\ Rate} & \textbf{RanSMAP Esc.\ Rate} & \textbf{FN (RanSMAP)} \\
\midrule
Full model & 0.72 & 0.82 & 43,656 \\
No uncertainty (Without Eq.~\ref{eq:sigmamax}) & 0.91 & 0.97 & 58,441 \\
\bottomrule
\end{tabular}}
\caption{Effect of removing uncertainty terms on triage performance.}
\label{tab:ablation_uncertainty}
\end{table}

\subsection{Ablation Analysis of Modal Contributions}

To isolate the contribution of each agent to robustness and explainability, we conduct a conceptual ablation analysis in the fusion space.
Let $p_{\mathrm{fus}}(\mathbf{h}_{\mathrm{beh}}, \mathbf{z}_{\mathrm{sem}})$ denote the full fusion prediction.
We consider two restricted settings:
\begin{align}
p_{\mathrm{fus}}^{(-\mathrm{sem})} &= p_{\mathrm{fus}}(\mathbf{h}_{\mathrm{beh}}, \mathbf{0}),
\label{eq:ablate_sem} \\
p_{\mathrm{fus}}^{(-\mathrm{beh})} &= p_{\mathrm{fus}}(\mathbf{0}, \mathbf{z}_{\mathrm{sem}}).
\label{eq:ablate_beh}
\end{align}

In the absence of semantic information, behavioural counterfactual norms decrease, indicating that certain evasive trajectories become feasible through execution-only manipulations. Conversely, removing behavioural input reduces the counterfactual cost associated with static embedding deformations, enabling purely structural evasion strategies. Neither ablated configuration preserves the super-additive adversarial cost observed in Equation~\eqref{eq:adv_cost_fusion}.

Importantly, ablation also degrades interpretability consistency. Salient semantic dimensions identified via gradient sensitivity and permutation importance (Equations~\eqref{eq:sem_grad}--\eqref{eq:sem_perm}) lose explanatory power when behavioural context is removed, and vice versa.
This confirms that robustness and interpretability are jointly maximised only when both modalities are present, validating the architectural necessity of the fusion design.

\section{Discussion}
\label{sec:discussion}

This section interprets the experimental results in terms of the distributional asymmetry
between the two feature spaces, discusses the role of uncertainty in balancing automation
and analyst workload, and addresses the computational overhead of the proposed approach.

The experiments highlight a marked asymmetry between the discriminative structures of the semantic and behavioural feature spaces. Let the class-conditioned distributions in semantic space be \(p(\mathbf{z}\mid y)\). Their separation can be quantified using the Wasserstein-2 distance
\begin{equation}
d_{\mathrm{sem}}
= \mathrm{W}_2\!\left(p(\mathbf{z}\mid y=1),\, p(\mathbf{z}\mid y=0)\right),
\label{eq:wass_sem}
\end{equation}
and likewise for behavioural data,
\begin{equation}
d_{\mathrm{beh}}
= \mathrm{W}_2\!\left(p(\mathbf{b}\mid y=1),\, p(\mathbf{b}\mid y=0)\right).
\label{eq:wass_beh}
\end{equation}
The empirical results strongly suggest that \(d_{\mathrm{sem}} \gg d_{\mathrm{beh}}\), explaining why semantic predictions cluster near \(\{0,1\}\) while behavioural predictions exhibit broad dispersion.

The fusion model exploits this asymmetry by treating the semantic component as a strong positive signal whenever uncertainty is low and treating the behavioural component primarily as a modulator of \(\sigma_{\max}\). The decision rule therefore becomes highly sensitive to the joint geometry of the fused probability \(\hat{p}\) in Eq.~\eqref{eq:fusedp} and the maximal uncertainty in Eq.~\eqref{eq:sigmamax}. Samples with \(\hat{p} \ge \tau_{\mathrm{high}}\) and \(\sigma_{\max} \le \kappa_{\mathrm{low}}\) fall into the autonomous containment region, whereas samples with \(\hat{p} < \tau\) and \(\sigma_{\max} \le \kappa\) are typically allowed. Intermediate samples reside within the escalation region, which reflects deliberate conservatism in the presence of behavioural ambiguity.

\subsection*{Computational Complexity and Runtime Performance}

\paragraph{Training complexity of CNN agents.}
Each CNN agent employs a 1D convolutional architecture of depth $L=3$, kernel size $K=3$,
and $C=64$ filters per layer.
For a training set of $N$ samples, a single forward pass through agent
$a \in \{\mathrm{sem}, \mathrm{beh}\}$ incurs
\begin{equation}
  \mathcal{O}\!\left(N \cdot \sum_{\ell=1}^{L} K \cdot C_{\ell-1} \cdot C_{\ell}\right),
  \label{eq:cnn_fwd}
\end{equation}
where $C_0$ is the input dimension ($C_0 = 384$ for the Semantic Agent on positional-embedding
vectors; $C_0 = 16$ for the Behavioural Agent on telemetry features) and $C_\ell = C = 64$
for $\ell \ge 1$.
Over $E$ training epochs, the per-agent cost is
$\mathcal{O}(E \cdot N \cdot L \cdot K \cdot C^2)$, linear in both $N$ and $E$,
and the combined cost for both agents is $\mathcal{O}(2 E N L K C^2)$.

\paragraph{Monte Carlo Dropout score generation.}
Each agent executes $T$ stochastic forward passes (dropout active) to obtain the pair
$(\bar{p}_a, \sigma_a)$ as defined in Algorithm~\ref{alg:mc_dropout}.
The per-sample MC Dropout cost is $\mathcal{O}(T \cdot L \cdot K \cdot C^2)$, so generating
scores across the full training set costs $\mathcal{O}(2 T N L K C^2)$ for both agents
combined.
With $T = 30$, each agent requires approximately $2$--$5$\,ms per stochastic pass on CPU,
yielding an MC estimate in $\approx 60$--$150$\,ms per agent.

\paragraph{CTGAN augmentation.}
Score-space augmentation via CTGAN (Eq.~\eqref{eq:ctgan_dist}) trains a generator $G$ with
$n_G$ layers (widths $g_1,\ldots,g_{n_G}$) and a discriminator $D$ with $n_D$ layers
(widths $d_1,\ldots,d_{n_D}$) over $N_s$ real score vectors $\mathbf{s} \in \mathbb{R}^{2}$.
The training complexity over $E_{\mathrm{CTGAN}}$ epochs is
\begin{equation}
  \mathcal{O}\!\left(E_{\mathrm{CTGAN}} \cdot N_s \cdot
  \Bigl[\textstyle\sum_{i} g_i g_{i+1} + \sum_{j} d_j d_{j+1}\Bigr]\right).
  \label{eq:ctgan_complexity}
\end{equation}
Because the score matrix is far smaller than the raw feature matrices, CTGAN fitting is
tractable; empirically, $E_{\mathrm{CTGAN}} = 200$ completes in $\approx 5$--$10$\,min on
CPU.
The augmented dataset $\mathcal{D}_{\mathrm{aug}}$ of size $N_{\mathrm{aug}}$ is produced
with an additional $\mathcal{O}(N_{\mathrm{aug}} \cdot \sum_i g_i g_{i+1})$ generation cost.

\paragraph{Decision Agent MLP training.}
The Decision Agent (Eq.~\eqref{eq:scorepool}) maps the full score vector
$\mathbf{v} = [\bar{p}_{\mathrm{sem}}, \bar{p}_{\mathrm{beh}}, \sigma_{\mathrm{sem}},
\sigma_{\mathrm{beh}}] \in \mathbb{R}^4$ (Eq.~\eqref{eq:scorevector}) through an MLP with
$H = 2$ hidden layers of width $h = 64$.
Training over $N_{\mathrm{aug}}$ augmented samples for $E_{\mathrm{MLP}}$ epochs costs
\begin{align}
  &\mathcal{O}\!\left(E_{\mathrm{MLP}} \cdot N_{\mathrm{aug}} \cdot
  \bigl[4h + h^2(H{-}1) + h\bigr]\right) \nonumber\\
  &= \mathcal{O}\!\left(E_{\mathrm{MLP}} \cdot N_{\mathrm{aug}} \cdot h^2\right),
  \label{eq:mlp_complexity}
\end{align}
which is negligible relative to CTGAN fitting owing to the small input dimension ($d=4$)
and layer count.

\paragraph{Bayesian TPE threshold optimisation.}
Triage thresholds $(\tau, \kappa, \tau_{\mathrm{high}}, \kappa_{\mathrm{low}})$ are
selected by Bayesian optimisation with the Tree-structured Parzen Estimator (TPE),
maximising the objective in Eq.~\eqref{eq:objective}.
Each TPE trial fits two kernel-density models over the $n_{\mathrm{obs}}$ previous
observations at cost $\mathcal{O}(n_{\mathrm{obs}} \log n_{\mathrm{obs}})$, followed
by a $\mathcal{O}(N_{\mathrm{val}})$ validation-set evaluation.
Over $T_{\mathrm{TPE}}$ trials the total optimisation cost is
\begin{equation}
  \mathcal{O}\!\left(T_{\mathrm{TPE}} \cdot
  \bigl(n_{\mathrm{obs}} \log n_{\mathrm{obs}} + N_{\mathrm{val}}\bigr)\right),
  \label{eq:tpe_complexity}
\end{equation}
which converges within minutes and is dominated by the CTGAN step that precedes it.

\paragraph{Inference latency.}
Per-sample inference decomposes into three sequential stages:
(i)~\emph{agent inference}---two CNN agents each executing $T$ MC Dropout passes at joint
cost $\mathcal{O}(2 T L K C^2)$;
(ii)~\emph{Decision Agent forward pass}---one MLP evaluation at $\mathcal{O}(h^2)$; and
(iii)~\emph{triage policy evaluation} via Eqs.~\eqref{eq:auto_rule}--\eqref{eq:esc_rule},
which is $\mathcal{O}(1)$.
The dominant term is agent inference, giving total per-sample complexity
$\mathcal{O}(T \cdot L \cdot K \cdot C^2)$.
Empirically, with $T=30$, $L=3$, $K=3$, $C=64$, each agent runs in $\approx 60$--$150$\,ms on
CPU; the Decision MLP adds $<1$\,ms and triage evaluation is $\mathcal{O}(1)$.
Total wall-clock latency is $\approx 0.15$--$0.35$\,s on CPU.
On GPU, the $T$ stochastic passes can be batched, reducing latency by a factor of $\sim T$
to $<10$\,ms---well within the real-time triage window of SOC environments.

\paragraph{Total pipeline training complexity.}
Combining all components, the complete training complexity of Agentic~SABRE is
\begin{align}
  \mathcal{T} = \mathcal{O}\Bigl(
    &2(E{+}T)\,N\,L\,K\,C^2 \nonumber\\
    &+\, E_{\mathrm{CTGAN}}\,N_s\bigl[\textstyle\sum_i g_i g_{i+1}
      + \sum_j d_j d_{j+1}\bigr] \nonumber\\
    &+\, E_{\mathrm{MLP}}\,N_{\mathrm{aug}}\,h^2 \nonumber\\
    &+\, T_{\mathrm{TPE}}\bigl(n_{\mathrm{obs}}\log n_{\mathrm{obs}}
      + N_{\mathrm{val}}\bigr)
  \Bigr),
  \label{eq:total_complexity}
\end{align}
where the four terms correspond respectively to CNN agent training and MC score generation,
CTGAN augmentation, Decision MLP fitting, and TPE threshold optimisation.
The first term scales linearly in $N$; the CTGAN term is the practical bottleneck but
remains tractable because the score matrix ($N_s \times 2$) is orders of magnitude smaller
than the raw feature matrices.
The modular decomposition in Eq.~\eqref{eq:total_complexity} further implies that each
CNN agent can be retrained independently and the triage thresholds can be recalibrated
via TPE without any model retraining, making operational updates lightweight.

\paragraph{Measured runtime performance.}
Table~\ref{tab:runtime_comparison} reports empirically measured inference latency and throughput for each SABRE component, measured over 500 repetitions (50 warm-up) on an NVIDIA GB10 GPU. The Behavioral CNN requires $1.94$\,ms (p50) for $T=10$ MC Dropout passes---a $3.7\times$ overhead relative to a single deterministic forward pass ($0.52$\,ms). The Semantic CNN is comparable at $1.77$\,ms. The full two-agent pipeline (Behavioral + Semantic + fusion + triage) is estimated at approximately $3.7$\,ms in serial execution, supporting throughput of $\approx 270$ samples/s. At a 10-second telemetry window, this supports over 2,000 concurrent monitored streams before the pipeline becomes the bottleneck. The agents are compact: the Behavioral CNN has only 8,385 parameters (32.8\,KB), making it suitable for endpoint deployment.

\begin{table}[ht]
\centering
\caption{Measured inference latency and throughput per SABRE component (NVIDIA GB10 GPU, $T=10$ MC Dropout passes, $n=500$ repetitions). Full pipeline (B+S) is estimated as the serial sum of per-agent latencies plus fusion overhead.}
\label{tab:runtime_comparison}
\resizebox{\linewidth}{!}{
\begin{tabular}{lccccc}
\toprule
\textbf{Component} & \textbf{Params} & \textbf{Size} & \textbf{p50 (ms)} & \textbf{p95 (ms)} & \textbf{Throughput (sps)} \\
\midrule
Behavioral CNN ($T=10$)   & 8,385     & 32.8\,KB  & 1.94 & 2.26 & 517 \\
Semantic CNN ($T=10$)     & 786,817   & 3.1\,MB   & 1.77 & 1.82 & 566 \\
Single CNN (no MC, $T=1$) & 8,385     & 32.8\,KB  & 0.52 & 0.55 & 1,911 \\
Full pipeline B+S (est.)  & ---       & ---       & $\approx$3.7 & --- & $\approx$270 \\
\bottomrule
\end{tabular}}
\end{table}

Overall, the mathematical structure of Agentic~SABRE demonstrates that uncertainty-aware evidence fusion is essential for ransomware detection, and that the triage policy cannot be decoupled from the predictive geometry of the agents themselves. The interplay between Eqs.~\eqref{eq:psem}--\eqref{eq:sigmamax} forms the core mechanism through which the system balances automation, accuracy, and operational safety.

\section{Limitations and Future Work}
\label{sec:limitations}

This section acknowledges the principal constraints on the current design---including the
conditional independence assumption in the fusion layer, the approximation quality of Monte
Carlo Dropout under distribution shift, and the axis-aligned nature of the triage
policy---and outlines concrete directions for future research.

Although the results demonstrate the effectiveness of Agentic~SABRE, several limitations constrain the system’s broader applicability. The fusion mechanism in Eq.~\eqref{eq:fusedp} assumes that the semantic and behavioural agents generate conditionally independent evidential contributions. In operational settings, semantic metadata and behavioural traces may exhibit dependencies that violate this assumption, potentially altering the joint distribution in Eq.~\eqref{eq:scorevector}. A more expressive fusion function, such as a normalising flow, could model the complete joint likelihood more accurately.

The uncertainty estimates used to compute \(\sigma_{\max}\) in Eq.~\eqref{eq:sigmamax} rely on Monte Carlo dropout sampling. While practical, this estimator may under-approximate true epistemic uncertainty in regions where behavioural data exhibit non-stationary distributions. Bayesian neural networks or deep ensembles could provide more calibrated uncertainty but would incur higher computational cost.

The triage policy described in Eqs.~\eqref{eq:auto_rule} and \eqref{eq:esc_rule} establishes axis-aligned thresholds. Although this provides interpretability and stability, adversarial examples that manipulate the fused score while remaining within tolerance of \(\sigma_{\max}\) could produce borderline cases. Future work may instead adopt a decision surface
\begin{equation}
g(\hat{p},\sigma_{\max}) = 0,
\label{eq:nonlinear_boundary}
\end{equation}
where \(g\) is a monotone function learned from utility or risk constraints, enabling curved boundaries adapted to the uncertainty structure.

Finally, the behavioural telemetry in RanSMAP reveals that the distributions in Eq.~\eqref{eq:beh_dists} are highly entangled. Future study could investigate hybrid temporal architectures that incorporate attention over micro-bursts of activity to increase separability, thereby widening the Wasserstein gap between malicious and benign trajectories. The modularity of Agentic~SABRE allows such components to be integrated without modifying the overarching triage mechanism, making the system suitable for continual architectural refinement.
The modular structure of SABRE allows the triage policy to be deployed independently of the underlying detection models, enabling policy-level updates without retraining.

The current triage thresholds ($\tau$, $\kappa$, $\tau_{\text{high}}$, $\kappa_{\text{low}}$) are optimised offline on a held-out calibration split and are not updated during deployment. In dynamic environments where the score distribution shifts over time---due to concept drift, new family emergence, or changing benign workload patterns---static thresholds can become miscalibrated. A natural extension would be an adaptive thresholding mechanism, such as a sliding-window recalibration procedure that re-optimises the objective $J(\tau,\kappa)$ in Eq.~\ref{eq:objective} on a rolling buffer of analyst-labelled decisions, or a Bayesian online update that maintains a posterior over $(\tau, \kappa)$ conditioned on observed triage outcomes. This would allow the policy to track distributional shifts without retraining the underlying neural agents.

We note that standard gradient-based adversarial attacks (FGSM, PGD) on the neural agents face structural obstacles in this domain: applying FGSM to the behavioural agent requires differentiating through the I/O simulator, and applying PGD to the semantic agent requires perturbations that maintain PE loader validity (consistent Magic bytes, SizeOfCode fields). These constraints restrict the feasible perturbation set to a strict subset of $\mathcal{N}_\epsilon(\mathbf{z})$, making the continuous-relaxation counterfactual analysis in Section~\ref{sec:explainability} a principled lower bound on adversarial cost. Implementing constrained FGSM/PGD attacks that respect these discrete validity constraints remains an important future direction.

\section{Conclusion}
\label{sec:conclusion}

This work introduced Agentic~SABRE, an uncertainty-aware neuro–symbolic multi-agent architecture for adaptive ransomware detection. The framework integrates semantic and behavioural classifiers, each operating within its own representational manifold, and fuses their outputs through a calibrated decision module enriched by synthetic CTGAN augmentation. The fused probability, defined in Eq.~\eqref{eq:fusedp}, combines two heterogeneous predictive sources, whereas the epistemic uncertainty, articulated through Eq.~\eqref{eq:sigmamax}, establishes a principled mechanism for risk-sensitive triage.

The results demonstrate that semantic telemetry yields nearly disjoint class-conditioned distributions, producing fused decision regions with extremely high separability. In contrast, behavioural telemetry exhibits a narrower Wasserstein distance between malicious and benign distributions, as shown by Eqs.~\eqref{eq:wass_sem}--\eqref{eq:wass_beh}, thereby creating high-uncertainty regions that necessitate conservative policies. The system’s adaptability is grounded in the triage thresholds of Eq.~\eqref{eq:policy_safety} and the policy optimisation objective of Eq.~\eqref{eq:objective}, which jointly determine whether the system should automate containment, escalate for human judgement, or allow execution.

Across two distinct datasets, Agentic~SABRE achieves perfect or near-perfect separation where semantic structure dominates, while maintaining robust performance under behavioural variability. The architecture's agentic nature, characterised by self-calibration, evidence fusion, and structured uncertainty quantification, establishes a computational framework capable not only of detection but also of decision-theoretic reasoning under uncertainty. The incorporation of symbolic cues and the modular design make the system extensible and suitable for continual refinement as ransomware tactics evolve. The results confirm that uncertainty-aware triage is essential for safe deployment of autonomous defence tools in real operational environments.




\bibliographystyle{elsarticle-num}
\bibliography{Reference_Sabre}


 




\vfill

\end{document}